\documentclass[a4paper,twoside, 11pt]{article}
\usepackage[normalem]{ulem}
\useunder{\uline}{\ul}{}
\usepackage{latexsym}
\usepackage{authblk}
\usepackage{amssymb}
\usepackage{amsmath}
\usepackage{amsthm}
\usepackage{booktabs}
\usepackage{enumitem}
\newcommand{\mycomment}[1]{}
\usepackage{graphicx}
\usepackage{subfig}
\usepackage{wrapfig}
\usepackage{color}
\usepackage{algorithm}
\usepackage[noend]{algpseudocode}
\usepackage{bm}
\usepackage{balance}
\usepackage{comment}
\usepackage{epsf}
\usepackage{svg}
\usepackage{tikz}
\usepackage{multirow}
\usepackage{placeins}
\usepackage{dashrule}
\usepackage{dashbox} 
\usepackage{hyperref}
\usepackage{cleveref}
\usepackage{natbib}
\DeclareMathOperator{\E}{\mathbb{E}}
\DeclareMathOperator{\I}{\mathbb{I}}

\usepackage[a4paper,
            left=1.2in,
            right=1.2in,
            top=1in,
            bottom=2in
            ]{geometry}
\hypersetup{hidelinks}

\providecommand{\keywords}[1]
{
  \small	
  \textbf{\textit{Keywords:}} #1
} 

\title{\Large \textbf{Efficient MCMC Sampling with Expensive-to-Compute and Irregular Likelihoods}}
\author[1]{\small Conor Rosato}
\author[2]{\small Harvinder Lehal}
\author[2]{\small Simon Maskell}
\author[2]{\small Lee Devlin}
\author[3]{\small Malcolm Strens}

\affil[1]{\footnotesize Department of Pharmacology and Therapeutics, University of Liverpool, UK;\newline \tt{cmrosa@liverpool.ac.uk}}
\affil[2]{\footnotesize Department of Electrical Engineering and Electronics, University of Liverpool, UK; \newline \tt{\{h.lehal, smaskell, ljdevlin\}@liverpool.ac.uk}}
\affil[3]{\footnotesize Independent Researcher, c/o Department of Electrical Engineering and Electronics, University of Liverpool, UK; \tt{malcolm@strens.com}}

\date{}
\usepackage{fancyhdr}
\begin{document}


\maketitle

\begin{abstract}

Bayesian inference with Markov Chain Monte Carlo (MCMC) is challenging when the likelihood function is irregular and expensive to compute. We explore several sampling algorithms that make use of subset evaluations to reduce computational overhead. We adapt the subset samplers for this setting where gradient information is not available or is unreliable. To achieve this, we introduce data-driven proxies in place of Taylor expansions and define a novel \emph{computation-cost aware} adaptive controller. We undertake an extensive evaluation for a challenging disease modelling task and a configurable task with similar irregularity in the likelihood surface. We find our improved version of Hierarchical Importance with Nested Training Samples (HINTS), with adaptive proposals and a data-driven proxy, obtains the best sampling error in a fixed computational budget. We conclude that subset evaluations can provide cheap and naturally-tempered exploration, while a data-driven proxy can pre-screen proposals successfully in explored regions of the state space. These two elements combine through hierarchical delayed acceptance to achieve efficient, exact sampling. 
\end{abstract}


\section{Introduction}
\label{sec:Introduction}


Sampling from a target density, such as a posterior in Bayesian inference of model parameters $\theta$, is often performed using Markov Chain Monte Carlo (MCMC) methods which originate in computational physics \citep{Metropolis1953}. MCMC can provide exact samples for target densities that are not well-approximated by a known parametric family. Other application domains where MCMC is applied to complex models include epidemiology \citep{moore2022refining}, genetics \citep{salehi2023cancer} and weather \& climate forecasting \citep{tomassini2007robust}. The target density usually factors across a prior and likelihoods of individual data points or subsets (batches).

When the parameters of the model are continuous and certain \textit{regularity} conditions are met, the gradient of the likelihood can be calculated and used to improve MCMC proposals. Gradient-based MCMC methods, including the Metropolis-adjusted Langevin Algorithm (MALA) \citep{roberts1996exponential}, Hamiltonian Monte Carlo (HMC) \citep{duane1987hybrid}, and the No-U-Turn Sampler (NUTS) \citep{NUTS}, exploit this gradient information to efficiently explore the target distribution. Gradients are used to obtain directional proposals, usually with the Metropolis-Hastings (MH) rule to account for their asymmetry.

\subsection{Irregular Likelihoods}

`Irregular likelihood' is an informal term for functions that do not meet common regularity conditions such as smoothness. For our purposes, it means that analytical gradients are either not available or are misleading (due to likelihood noise, or chaotic regimes, for example). This excludes the use of fast gradient-based MCMC, and so simple random walk proposals are often used instead. These are typically much slower to explore the target density than the gradient-based methods, particularly in high-dimensional state spaces, potentially leading to poor-quality samples. Our work is also relevant for non-Euclidean state spaces, for example, with ordinal (ordered categorical) dimensions or even variable dimensionality.

The sampling challenge is greatest when the likelihood function is very expensive to compute. This arises when a simulation or filtering process must be run to obtain the likelihood for each item in the data set. A `tall' dataset (containing many examples) scales the computational burden even more. 

One example which can have these undesirable characteristics is Approximate Bayesian Computation (ABC) \citep{beaumont2002approximate}, where the likelihood is intractable and replaced by a distribution obtained by matching repeated-simulation outputs to the real data, using a distance metric.

Another example, which is our main focus, is calibrating stochastic dynamic systems \citep{barber2011bayesian}, such as epidemiological state space models. Specifically, when modelling new infectious diseases, it is necessary to infer the parameters of the model and to understand their uncertainty: knowing the infection and recovery rates of a particular disease allows public health officials and researchers to ascertain the reproductive number `R' which quantifies the average number of secondary infections produced by a single infected person. 

Observed time series can be processed by a particle filter (PF) to compute a likelihood \citep{calvetti2021bayesian, endo2019introduction, rasmussen2011inference} given specific model parameters. The PF likelihood is used in particle-MCMC (PMCMC) \citep{andrieu_doucet_holenstein_2010}, a Sequential Monte Carlo (SMC) method that is particularly challenging in terms of computation. PMCMC is widely used for Bayesian inference where direct likelihood evaluation is infeasible due to the stochastic transitions of latent states.  

In normal operation, a PF processes a time series to infer the latent states of a dynamic model \citep{arulampalam2002tutorial, elfring2021particle}. In PMCMC, the PF \textit{also} provides an unbiased, but noisy, estimate of the intractable likelihood of the observed time series \emph{conditional on fixed parameters of the stochastic data-generating process}. This `inner' PF inference process can therefore be embedded in an `outer' MCMC chain over the parameters. 


If sufficient computation resource is available to evaluate likelihood for the whole data set at once, a pseudo-marginal MCMC chain can use the (noisy but unbiased) PF likelihoods directly and still guarantee samples from the correct posterior. However, computational demand can be prohibitive since each full likelihood evaluation requires a PF run for every data item (time series), and within each such run there are many parallel computations (particle updates). Therefore, it is often necessary to approximate the full likelihood using a structured approach. 

One way to do this is by dividing the computation into smaller, manageable chunks, which we refer to as `scenarios'. In this work, scenarios divide the data, but in \Cref{sec:discussion} we discuss alternatives. The decomposition of the data allows for partial likelihood evaluations, where only a \emph{subset} is processed at a time. Existing `subset MCMC' methods aim to make best use of these partial likelihoods, exploiting known bounds or statistical tests, but have not been explored for irregular likelihoods.  We now formalise the decomposition of the full likelihood and provide an illustrative example before outlining our contribution.

\subsection{Scenarios \& Subsets}

Let a data set be divided into $N$ parts, $(x_i)_{i=1}^N$, indexed by scenario $i$. Assuming independence, the target density $\pi(\theta)$ for MCMC sampling factors across scenarios:
\begin{align}
\pi(\theta) \propto p(\theta) \prod_{i=1}^N p(x_i\mid\theta),
\label{eq:target}
\end{align}
where $p(\theta)$ is an optional prior density.
A single scenario evaluation involves processing the associated chunk of data to obtain its likelihood (up to an unknown normalising constant):
\begin{align*}
L_i(\theta) \propto p(x_i|\theta).
\end{align*}

\textit{Within} each scenario likelihood evaluation the data items need not be independent; \textit{e.g.,} they can take the form of a time series. In this work, we divide our data sets into $N=64$ scenarios, each containing a batch or time series. The product of likelihoods in \Cref{eq:target} leads to additive structure in \emph{log} likelihood, so it is convenient to also define:
\begin{align*}
l_i(\theta) = \log L_i(\theta).
\end{align*}

When there is an embedded simulation (perhaps repeated over multiple random seeds) or a PF run within each scenario likelihood evaluation, the product in \Cref{eq:target} can be very expensive to compute even when the data set is relatively small. With small data sets, more irregular (non-Normal) target densities are obtained because asymptotic convergence results, discussed further in \Cref{subsec:proxies}, are not applicable. 

Consider that we have a limited computational budget expressed as a number of single-scenario likelihood evaluations. Likelihood evaluations on subsets of the $N$ scenarios can be beneficial in two ways:
\begin{itemize}[noitemsep,topsep=0pt]
    \item When subset evaluations are used in the top-level sampling chain, more iterations can be obtained (and thus a larger sample) within the computational budget.
    \item When subset evaluations are used to pre-screen proposals, or to obtain directed proposals, for a full evaluation, this can lead to larger steps being taken or higher acceptance rates, and so more effective exploration of the target density.
\end{itemize}

\subsection{Motivating Example}
\label{subsec:Motivation}

Suppose that each data point $x_i$ in a data set is a noisy observation dependent on a hidden (latent) state $z_i$ of a stochastic dynamic system at some point in time. The latent state is dependent on a model parameter $\theta \in \mathbb{R}$, which is the target for inference. Let observed data $(x_i)_{i=1}^N$ be Poisson-distributed counts with means $z_i/100$.

%
The distribution of $z_i$ is not usually available in analytical form, but samples can be drawn by simulation of a stochastic system for multiple time steps. For illustration only, let us replace this stochastic simulation with a single Poisson draw\footnote{When drawing samples we assume the random number source may be interpreted differently for evaluations at different states unless they fall within the same cell on a high-resolution grid in $\theta$. This reflects the reality of stochastic dynamic simulations with many time steps.}, $z_i \sim \text{Poisson}\left(10^5 S(\theta)\right)$ where a sigmoidal transformation $S$ is applied to state $\theta$ to ensure the parameter is positive.
%
To obtain the likelihood $p(x_i\mid\theta)$ we would need to integrate over $z_i$, requiring an infinite number of simulation runs. Taking the average of $p(x_i|z_i^j, \theta)$ over a fixed finite sample $\{z_i^j\}_{j=1}^m$ provides an estimate for this intractable likelihood.
\begin{figure}
    \centering
    \includegraphics[width=0.9\linewidth]{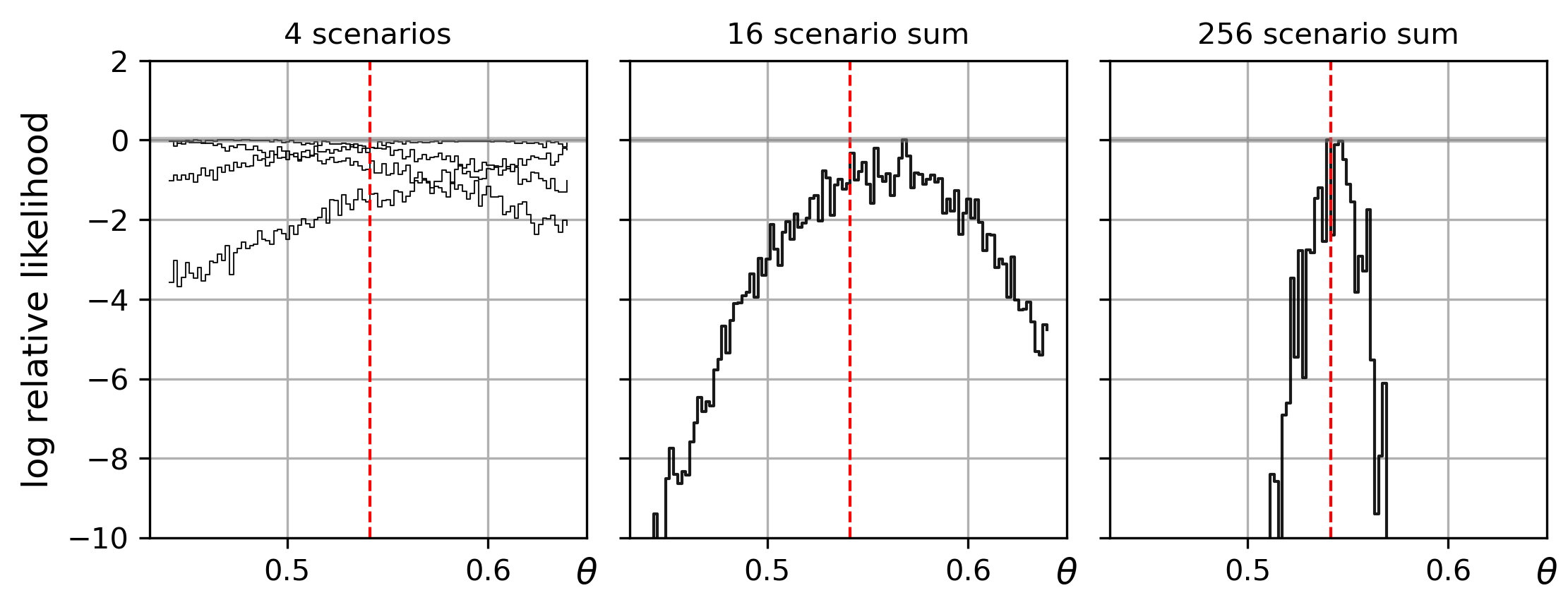}
    \caption{Simplified example of an irregular log likelihood.}
    \label{fig:motivating example}
\end{figure}

For context, in a disease model such as the one presented in Appendix \ref{app:time_series_PF_example}, the latent state is the unknown infection rate in a population, the observations are counts from testing individuals, and the state space for sampling encodes characteristics of the disease like reproduction rate. Furthermore, observations are time series from a dynamic model with latent state, hence requiring an expensive PF likelihood evaluation. The $m$ parameter (number of random draws) above is analogous to the number of particles. 

For illustration, consider the single-sample case, $m=1$. When sufficient of these noisy log likelihoods are summed \emph{over data examples}, we will see the overall shape of the likelihood surface emerge. In \Cref{fig:motivating example}, we have simulated $N=256$ observations with $\theta = 0.541$, and computed log likelihood functions for each over a range of $\theta$ that might be explored by a sampler. Shown on the left are individual scenario log likelihoods $l_1(\theta),\ldots,l_4(\theta)$, displaced so that their maxima are at zero on the y-axis. These functions individually exhibit noise that varies across small changes in $\theta$, and gradients would not be useful. Summing the log likelihood over 16 observations (middle) shows a clear mode, slightly displaced from the true $\theta$ (vertical dashed line). Therefore, this subset contains useful information about the target in \Cref{eq:target}. Finally, the full likelihood function (on the right) has a mode that shows very little bias. Placing these functions on the same scale highlights that the variance of the noise \emph{increases} with more data, and so we need sampling algorithms that are robust to this noise even in the large data limit. For a multi-dimensional version of this task, see \Cref{app:synthetic}. 

\subsection{Subset Samplers}\label{subsec:subset_samplers}

In this paper, we take a number of existing samplers and adapt them to operate for irregular likelihoods, as well as integrating acceleration methods (proxies and adaptive proposals):
\begin{itemize}[itemsep=1pt,topsep=1pt]
    \item Baseline: full MCMC with random walk proposals in which the full sum of scenario likelihoods is evaluated on every iteration.
    \item Subsample MCMC \citep[\textit{e.g.,}][]{Bardenet2017} is an approximate method that uses subsets directly in the top-level chain. The subset log likelihood is rescaled so that it has the same magnitude as the full likelihood.
    \item The Austerity sampler \citep{Korattikara2014} is similar to Subsample MCMC but increases the subset size until a statistical test indicates confidence that the proposed state would be accepted or rejected on a full evaluation. 
    \item The Confidence sampler \citep{Bardenet2017} employs a more conservative statistical test with fewer assumptions regarding the distribution of single-scenario likelihoods. It can improve subset evaluations by incorporating a cheap-to-compute approximation (a `proxy') to reduce variance and generate better samples.  
    \item The Firefly sampler \citep{maclaurin2014} uses a strong lower bound to obtain a factored form of the likelihood. By introducing auxiliary binary state variables that activate and deactivate scenarios, it allows exact sampling while relying on subset evaluations. 
    \item The HINTS sampler \citep{Strens2004} runs a top-level chain that uses full evaluations, but constructs directed proposals for this chain using subset evaluations, organised in a hierarchy. Using the known asymmetry of the subset-based proposals, the top-level chain generates exact samples. Typically, the steps taken are much larger than for full MCMC with undirected random walk proposals. 
\end{itemize}
In comparative evaluations, we emphasise the need for minimal task-specific tuning of parameters and like-for-like comparisons by using common components such as proxies. Source code to replicate all results in this work is available at: \url{https://anonymous.4open.science/r/SubsetSamplers-3142/}.


\subsection{Contribution \& Paper Outline}

\Cref{sec:subset_MCMC} provides an overview of adaptive MCMC and the proxy functions used in subset samplers. Each of the existing subset samplers introduced above are reviewed, with distinctions drawn between approximate and exact approaches.
 
Our novel algorithmic contributions are presented in \Cref{sec:improvements}. 
The proxy functions used by subset samplers are often obtained by Taylor expansions, which rely on analytical derivatives, assumed here to be \emph{unavailable}. In \Cref{subsec:dd_proxy} we propose a \textit{quadratic} data-driven proxy that is consistent with theoretical predictions for the log likelihood's shape. We also introduce a nearest-neighbour proxy with a distance cut-off, for non-quadratic tasks. Each data-driven proxy is fitted on-line to the sample history, including rejected proposals. We show for the first time how a proxy can be integrated into HINTS. 

With expensive likelihood evaluations exceeding the overhead of running the samplers, the computation per iteration of a subset sampler varies according to the number of evaluations needed to pass a statistical test or to construct a directed proposal. This means that better samples should be obtained if the optimisation objective for adapting proposals is the progress made by the chain \textit{per unit of computation}, instead of progress per iteration. \Cref{subsec:computation_adaptive} proposes a new algorithm for \emph{computation-cost aware} adaptive MCMC. We show that the new algorithm extracts good performance from all the samplers while avoiding a preliminary search over proposal scale parameters. Of particular note is that some samplers operate in an unexpected regime where they use a very low acceptance rate but make very large moves in state space. In \Cref{subsec:modified_sampler_testing} we verify that each of the samplers operates effectively on an `easy' (noise-free) task and explore how the new adaptive control and data-driven proxies affect them. 

A theoretical contribution is made in \Cref{app:proof} where we prove, for the first time, that HINTS is an exact sampler. The result applies not just to the original (additive log likelihood) formulation but to a broader scheme where there is flexibility in the functions that extract information from subsets, including proxies.

For approximate samplers, it is essential to make comparative evaluations to measure accuracy. Even for exact samplers, experimental evaluation is valuable because excessive mixing times can damage accuracy. \Cref{sec:evaluation} provides a number of large-scale comparisons between the improved samplers on a range of challenging `irregular likelihood' tasks, including calibration of a disease model using PMCMC-style likelihoods. Using the same cost-aware adaptive control and data-driven proxies (where applicable) helps to ensure like-for-like comparison of samplers. We chose tasks that allow us to measure actual sample error in addition to the usual mixing metrics.

Beyond the disease model calibration, we perform comparisons with several challenging task variations: \emph{high correlation} between state dimensions, \emph{higher dimensional} states, and \emph{tall data}. On the more challenging tasks, we find that some subset samplers either confer no computational advantage (over full MCMC) or pay a price in terms of accuracy; with the exception of HINTS (with proxy).

\Cref{sec:discussion} offers broader discussion and identifies connections with related work and recent developments. \Cref{sec:Conclusions} concludes and identifies future research directions.





%
%
%
%

\section{Subset Metropolis-Hastings MCMC}
\label{sec:subset_MCMC}

MCMC sampling simulates a Markov chain from a target density $\pi(\theta)$ such that the output states are a true sample. The MH algorithm \citep{Hastings1970} supports asymmetric proposal densities (such as gradient-based proposals), written $q(\theta'|\theta)$. The acceptance probability for proposed state $\theta'$ from current state $\theta$ under the MH Rule is 
\begin{align}
\alpha(\theta, \theta') =  1 \wedge \frac{\pi(\theta') q(\theta | \theta')}{\pi(\theta) q(\theta' | \theta)},
\label{eq:MH1}
\end{align}
where $a \wedge b = \min(a,b)$. If accepted, the proposal is taken as the new state of the chain; otherwise, the old state $\theta$ is retained. 


Often there is an unknown normalising constant ($Z$) on the target density: let $F(\theta) = \pi(\theta)/Z$. Then each iteration of \Cref{alg:MH} is an application of \Cref{eq:MH1}. A start state is specified, and the algorithm is iterated indefinitely with each $\theta_{out}$ used as the next $\theta_{in}$. After a ``burn-in" period (which lasts until the chain has ``mixed"), the samples generated will be consistent with the target density, when used with a proposal that ensures ergodicity and aperiodicity. 

\begin{algorithm}
\caption{Metropolis-Hastings ($\text{mh\_step}$)}\label{alg:MH}
\begin{algorithmic}[1]
\State Input: $\theta_{in}$ current chain state
\vspace{0.25em}
\State $\theta'\sim q\left(\theta'|\theta_{in}\right)$ \Comment{Propose new state}
\vspace{0.25em}
\State $\alpha = 1 \wedge \frac{F(\theta')}{F(\theta_{in})}\frac{q(\theta_{in}|\theta')}{q(\theta'|\theta_{in})} $ \Comment{Acceptance probability}
\vspace{0.25em}
\If{$\alpha > u \textrm{ with } u\sim\mathbb{U}[0,1]$} 
    \State Output: $\theta_{out}\gets \theta'$ \Comment{Accept}
\Else
    \State Output: $\theta_{out}\gets \theta_{in}$ \Comment{Reject}
\EndIf
\end{algorithmic}
\end{algorithm}

The simplest example of a proposal distribution in a vector space $\mathbb{R}^d$ is a Gaussian random walk proposal $q(\theta'|\theta)=\mathcal{N}(\theta,\Sigma)$ where $\Sigma$ is the covariance of the proposal. If the state space has been adequately `preconditioned' (\textit{e.g.,} scaled by $\Sigma^{-1/2}$) then we can work with:
\begin{align}
    q(\theta'|\theta)=\mathcal{N}(\theta,\sigma \I_D),
\label{eq:rw}
\end{align}
where $\sigma$ is a single task-specific scale parameter and $\I_D$ is the unit matrix of dimension $D$. This assumes states are unbounded, which is often achieved by transformation of individual dimensions; for example $\mathbb{R^+}$ is mapped to $\mathbb{R}$ by a logarithmic or sigmoidal function.


\subsection{Adaptive MCMC}\label{subsec:adaptive_MCMC}

The optimal $\sigma$ for simple random walk proposals depends on the task characteristics such as the typical change in $F(\theta)$ over a certain Euclidean distance. In general, this is unknown, and so a wide range of values are sometimes explored before the main sampling run. However, the optimal value will be different in the initial `burn-in' stage (where large changes in likelihood are often encountered) than the later `mixing' stage in which a final sample is generated. In general, it is not possible to establish when burn-in is complete. Therefore, finding a value for $\sigma$ by performing only a preliminary search may be unsuccessful.

Adaptive MCMC methods aim to find suitable proposal scale parameters automatically and converge to a value of $\sigma$ that yields good mixing performance. A wide range of adaptive MCMC methods are available for the random walk proposals used here \citep{Roberts01012009}. Some will scale the proposal density equally in all state dimensions, while others adapt the full covariance matrix of proposals (in a vector space). To limit complexity, we assume that best efforts have been made to precondition the state space so the adaptation can focus on a single scalar `multiplier', $r$. For the random walk proposal in \Cref{eq:rw} we have $\sigma = r\sigma_0$, where $\sigma_0$ is a suitable default for the task.

The simplest approaches adapt the proposal scale to control the average acceptance probability of the sampling chain. The target acceptance probability is often determined using a heuristic rule based on state space dimensionality. In \Cref{subsec:computation_adaptive} we explore alternative approaches for adaptive MCMC with subset samplers and irregular likelihoods.

Adaptation must be applied carefully because, if the proposal $q(\cdot)$ changes between evaluating the numerator and denominator of \Cref{eq:MH1}, the reversibility property is not guaranteed, and a true sample may not be obtained. This is usually mitigated by reducing the adaptation rate over time, or freezing $\sigma$ for the final sampling interval. We refer to samples obtained with a fixed $\sigma$ as `strict'. 

\subsection{Proxies in Subset Samplers}
\label{subsec:proxies}

Control variates are widely employed in MCMC and machine learning to reduce the variance of estimated quantities. Proxies---also known as surrogate functions---are cheap-to-compute expansions or approximations of the likelihood that can act as control variates for MCMC. Most subset samplers can benefit from variance reduction on the \emph{`subsampling noise'}: the differences between subset-based estimates and full evaluations of the target. This is usually achieved by incorporating a proxy function into the acceptance decision.

Proxies can be simple functions of state or specific to scenarios (individual data items or batches). Scenario-specific proxies can be more effective because `actual' likelihood evaluations on a subset can be combined with proxy evaluations for the remaining scenarios. However, the exact way in which proxies are used varies considerably between samplers.

The Bernstein-von Mises (BvM) theorem states that under certain regularity conditions, the posterior distribution $\pi(\theta)$ of model parameter vector $\theta$ tends to a Multivariate Normal distribution (MVN) as the data set size ($n$) grows:
\begin{equation}
\pi(\theta \mid X_1, X_2, \ldots, X_n) \approx \mathcal{N}\left(\hat\theta, \frac{I(\theta_0)^{-1}}{n}\right),
\label{eq:BVM}
\end{equation}
where $X_1, X_2, \ldots, X_n$ are the independent and identically distributed data. The limiting distribution is located at the maximum likelihood\footnote{A suitable prior can be included, in which case $\hat{\theta}$ is the max a posteriori (MAP) state.} state $\hat\theta$ and its covariance is the inverse of the Fisher information matrix $I(\theta_0)$ at the true state $\theta_0$ \citep{Mises1937WahrscheinlichkeitSU}.

If all the conditions for BvM were satisfied, we could compute the distribution and draw samples directly rather than run a sampling chain. Most Bayesian inference tasks of interest do not meet the BvM conditions because the data is finite or the model is misspecified (being a simplification of the true data generating process). The approximation becomes worse for state space models (like the disease model) as a result of several common characteristics: irregular likelihoods; correlations in the target density (multiple explanations for the same data) and small data counts.  A large body of research exists on the circumstances in which BvM is reliable \citep{BVMmiss, Bochkina_2023}.

The log likelihood of the MVN is a quadratic function. Even though BvM does not hold for our tasks, it motivates us to try a global quadratic proxy; we will see this can be useful in some samplers \textit{even when it is a poor approximation}.  Furthermore, once a sampler has `burnt-in' and discovered a region of high likelihood, the \textit{local} (\textit{i.e.,} sample-weighted) quadratic fit may be accurate for the important region of a non-quadratic target. Per-scenario (or subset) log likelihoods are much less likely to approach the quadratic form than the total log likelihood.

Proxies obtained by Taylor expansion of the log likelihood are often employed in subset samplers. These are also referred to as Parameter Expanded Control Variates in \citet{quiroz2018speeding} and make use of analytical derivatives. Inaccuracies are often mitigated by updating the proxies from time to time during sampling (`dropped proxies'), usually at the expense of reversibility (exact sampling) guarantees. Alternatively, where a quadratic proxy fails, more flexible structures can be introduced on a problem-specific basis, exploiting known structure in the likelihood where possible. 

\subsection{Subset Samplers}
\label{sec:subsetSamplers}

We now introduce each of the subset-based samplers to be used in our experimental comparison. \Cref{tab:original} shows the list of samplers we work with and the extent to which they have been evaluated with proxies (or bounds) \emph{to aid subsampling} and adaptive proposals. (For comparison,  \Cref{tab:improved} will show the changes we have made for this work.)

\begin{table}[h]
\centering
\begin{tabular}{lccc}
\toprule
\textbf{Sampler}& \textbf{Exact/Approximate} & \textbf{Proxy} & \textbf{Adaptive} \\
\midrule 
Full MCMC & Exact & N/A & Yes \\
Subsample MCMC & Approximate & No & No \\
Austerity & Approximate & No & No \\
Confidence & Approximate & Analytical & No                               \\
Firefly & Exact & Analytical bound& No                               \\
HINTS& Exact (without proof) & No & No                               \\     
\bottomrule
\end{tabular}
\caption{Samplers \& use of subset-noise reduction proxies or adaptive proposals.}
\label{tab:original}
\end{table}

\subsubsection{Subsample MCMC} 

We include a `Subsample MCMC' sampler to illustrate the impact of using random subset evaluations in place of full likelihood evaluations in an MCMC chain. The total log likelihood is approximated by the rescaled evaluation for a scenario subset $S$ with fixed size $|S|$, randomly drawn without replacement at every step:

$$\sum_{i=1}^N l_i(\theta) \approx \frac{N}{|S|} \sum_{i \in S} l_i(\theta).$$ 

The potential speed-up is $N/|S|$ but there is no guarantee that the sample will be accurate. This naive approach can only be operated with a small, fixed step size because an adaptive step size could lead to arbitrarily poor samples.

\subsubsection{Austerity Sampler}

In Austerity Metropolis-Hastings \citep{Korattikara2014} instead of using a fixed-size subset, each accept or reject decision can be made on a subset if confidence is obtained, from a statistical test, that the same decision would be made on the full set. If not, the subset size is progressively increased, until the whole set of scenarios is considered, if necessary.

For the full set of $N$ scenarios, consider the mean log likelihood ratio per scenario:
\begin{equation}
\label{eq:Lambda}
    \Lambda_{N}(\theta,\theta')=\frac{1}{N} \sum^N_{i=1}\log\frac{p(x_i|\theta')}{p(x_i|\theta)} = \frac{1}{N} \sum^N_{i=1} l_i(\theta') - l_i(\theta).
\end{equation}

A full application of MH compares $\Lambda_{N}(\theta,\theta')$ with a threshold that depends on the prior ratio, the proposal ratio, and a uniform random sample, according to \Cref{alg:MH}. Let this threshold be $\mu_0$. The equivalent of \Cref{eq:Lambda} using a subset of only $b$ scenarios (randomly drawn without replacement) is written $\Lambda_{b}^*(\theta,\theta')$. Using the sample standard deviation $s_b$ of the subset, the standard deviation of the sample mean can be estimated as $\hat{s}=\frac{s_b}{\sqrt{b}}\sqrt{1-\frac{b-1}{N-1}}$ and a test statistic constructed:

$$\mathfrak{t} = \frac{\Lambda_{b}^*(\theta,\theta')-\mu_0}{\hat{s}}.$$

A 2-tailed Student-t test with $b-1$ degrees of freedom is applied to determine whether the accept/reject decision (according to the sign on $\mathfrak{t}$) can be made with the subset; otherwise the subset size increases geometrically (always reusing the existing evaluations), stopping when the test passes or an evaluation is made on all $N$ scenarios.

Whilst the Austerity sampler reduces the average number of likelihood evaluations per iteration, the t-test relies on the assumption that the subset means are Normally distributed. A Central Limit Theorem argument can be made for this (with very large $b$), but often it will not hold and inaccuracy was demonstrated in \cite{Bardenet2017} when fitting even a simple one-dimensional Lognormal density. Nevertheless, the strong burn-in performance (finding high-density areas of likelihood) makes the Austerity sampler worth including.

In our evaluations, we operated the Austerity t-test with a confidence level of 98\% (2 tailed), a minimum subset size of $N/16$ scenarios, and a subset geometric growth factor of $3/2$. Therefore, subsets of size 4, 6, 9, ...,  capped at $N$ were used.

\subsubsection{Confidence Samplers}

\cite{pmlr-v32-bardenet14} introduced a Confidence sampler that, like Austerity, applies a statistical test to the accept/reject decision, growing the random subset of scenarios until the test reaches confidence or the full set is evaluated. The aim is to bound the probability $\delta$ of an error that exceeds some $c>0$ between the estimate (obtained with subset of size $b$) and the full evaluation:

$$\mathbb{P}\left(|\Lambda(\theta,\theta')-\Lambda_b^*(\theta,\theta') \leq c|\right) \geq 1 -\delta.$$

In establishing this bound from statistics of the individual scenario likelihoods, the Confidence sampler makes weaker assumptions about the distribution of the individual scenario likelihoods than Austerity. Our implementation follows that of \cite{Bardenet2017} and uses a Bernstein concentration bound that makes use of the mean, maximum, and minimum of the likelihood differences (for the active subset).  

For well-understood likelihoods (with bounded gradients) it is possible to obtain some theoretical bounds on the quality of the sample that is generated. In the case of our irregular likelihoods these bounds will not hold; however, the stronger test (compared with Austerity samplers) is likely to make more cautious decisions, reducing the risk of large errors (\textit{e.g.,} visiting states that have arbitrarily low likelihood). 

An improved Confidence sampler was presented in \cite{Bardenet2017}, which introduces proxies. This works by `correcting' the quantity  $\Lambda_{b}^*(\theta,\theta')$ by adding the cheap proxy evaluation for the scenarios that are not in the subset. The goal for the inclusion of the proxy is that most of the subsampling error is cancelled. Quadratic proxies based on local Taylor expansions were evaluated. Furthermore, regularly updating the proxy helped mitigate errors, even though these `dropped' proxies break reversibility.

We ran the Confidence sampler with the same subset size sequence as for the Austerity sampler. For irregular likelihoods, we had to use a large tolerance ($\delta=0.1$) to obtain any significant speed-up over full MCMC.

\subsubsection{Firefly Sampler}

The Austerity and Confidence samplers are not `exact' in the sense that they can output states that are arbitrarily improbable under the full likelihood, for tasks where there is no known bound on the likelihood gradient. We now consider two samplers that can provide exact samples for \textit{irregular} likelihoods. 


The Firefly sampler, proposed by \cite{maclaurin2014}, augments the state space with a binary auxiliary variable, $z_i$ for each scenario (data point or batch), $x_i$. A strictly positive, cheap-to-compute lower bound $B_i(\theta)$, to the likelihood $L_i(\theta)$ is assumed to be known. Each auxiliary variable is given the Bernoulli distribution:

$$ p(z_i|x_i,\theta)=\left[ \frac{L_i(\theta)-B_i(\theta)}{L_i(\theta)}\right]^{z_i}\left[ \frac{B_i(\theta)}{L_i(\theta)}\right]^{1-z_i}.$$

Scaling by the denominator, we obtain:

$$ L_i(\theta)p(z_i|x_i,\theta)=\left[L_i(\theta)-B_i(\theta)\right]^{z_i}\left[B_i(\theta)\right]^{1-z_i}.$$

The key point is that the right-hand side only requires the actual (expensive) likelihood to be calculated for scenarios where $z_i = 1$. The remaining scenarios are `dark' until the auxiliary variables are resampled. This means that exact samples can be obtained from the augmented target density on $(\theta, (z_i)_{i=1}^n)$ with only subset evaluations. Discarding the auxiliary variables provides a true sample for the total likelihood (or the posterior if a prior is present).

Sampling proceeds by repeating the following process: resample (some or all of) the auxiliary variables (to create a new subset of `lit' scenarios), then run a regular MCMC chain for one or more steps using that subset. A new state is added to the sample at every step. Therefore, each state output is obtained with a fraction of the computation needed for conventional MCMC that requires a full likelihood evaluation on all scenarios at every step. Firefly can perform poorly if the lower bound is not sufficiently tight because few scenarios will be dark and there will be little computational benefit over full MCMC. Furthermore, the mixing time of the Firefly chain can be longer than full MCMC, depending on factors like how frequently the auxiliary states are resampled. 

Firefly makes use of a \emph{lower bound} rather than an approximating function. This lower bound would usually be derived on a task-specific basis, but we want to include Firefly in our evaluation with a generic functional form for the lower bound. BvM again motivates using a quadratic form for this bound. 

\subsubsection{Hierarchical Importance With Nested Training Samples (HINTS)}\label{subsubsec:HINTS}


Bayesian inference was one of the applications conceived for the HINTS sampling scheme proposed in \cite{Strens2004}. The main idea is to use `cheap' evaluations on data subsets to build directed proposals for the costly evaluation on the full data set; \textit{i.e.,} a successful move on a subset provides a `hint' for a more expensive evaluation. We interpret HINTS as a recursive application of \emph{delayed acceptance} \citep[\textit{e.g.,}][]{Christen01122005}. In delayed acceptance, a cheap-to-compute approximation $\pi^*(\cdot)$ of the target density is available. Only proposals that are first accepted by the MH rule with $\pi^*(\cdot)$ substituted for $\pi(\cdot)$ are allowed into a full MH rule that uses (expensive) evaluations of $\pi(\cdot)$. 

In addition to quantities used previously ($\theta$, $L_i(\theta)$, $N$), we define:
\begin{itemize}[itemsep=0pt, topsep=4pt]
\item $h \in \{0,\ldots,H\}$ is the level in the hierarchy; $h=0$ for leaf nodes and $H$ at the root.
\item $(m_h)_{h=0}^H$ are the branch factors (number of possible child nodes) at each level, except $m_0$ is the number of scenarios in each subset at leaf nodes. Therefore, this sequence determines the sampling hierarchy (or `tree structure').
\item $(n_h)_{h=0}^H$ are the number of nodes at each level, given by the product of the branching factors at higher levels: $n_h = \prod_{h'>h} m_{h'}$ and $n_H=1$ for the unique root-node.
\item The whole data set consisting of $N$ scenarios is associated with the root-node. At level $h$ the data is partitioned into $n_h$ subsets (one per node) each of size $N/n_h$. We denote these subsets $\mathbb{X}_j^h$ for $j=1,\ldots,n_h$.
\item $\text{child}(h,j,k)$ is the node index at level $h-1$ for the $k$\textsuperscript{th} child (of node $j$ at level $h$).
\item Scalar $\Psi$ is the asymmetry (Hastings' correction) associated with each  proposal, and takes the place of $\frac{q(\theta|\theta')}{q(\theta'|\theta)}$ in the MH rule of \Cref{eq:MH1}. For composite proposals it is accumulated as a product from multiple lower level steps.
\end{itemize}

We can now introduce a subset information function, $F^h_j(\theta)$, which is used to evaluate a state $\theta$ against the subset $\mathbb{X}_j^h$. In our generalised view of HINTS, the user has freedom to decide on a function that they expect will be informative for comparing states using the subset, except at the root-node we require $F^H_1(\theta) \propto \pi(\theta)$. Below the root, the obvious choice is the likelihood of the subset: $F^h_j(\theta) = \prod_{i \in \mathbb{X}_j^h}L_i(\theta)$. Likelihoods calculated in lower-level evaluations (sub-subsets) can be reused through caching at the scenario level. A prior $p(\theta)$ can optionally be included by scaling each term in this product by $p(\theta)^{1/n_h}$. This means the prior is properly apportioned across subsets at each level, and the whole prior is included at the root.

\Cref{alg:HINTS} defines how HINTS makes a decision. This function is called once per step of the top-level sampling chain, which operates at the root-node ($h=H$). Except at leaf nodes ($h=0$), execution proceeds by running a chain for $m_h$ steps where each step visits a child node (recursively), so a depth-first traversal of the hierarchy is being followed.

When execution first reaches a child node, a simple proposal is accepted or rejected using a leaf subset (which could be a single scenario). The resulting $\theta_{out}$ is passed back up the hierarchy, along with its asymmetry measure. This becomes an asymmetric proposal for the first step in the chain at the parent node, followed by further steps with subsequent leaves. In this way, bigger composite proposals are formed and returned back up the tree, each carrying a Hastings' correction $\Psi$, accumulated as a product from lower level chains, to track their asymmetry. Once all nodes have been traversed, a highly-directed proposal built from many simple moves will be accepted or rejected at the root-node. The only retained state is $\theta_{out}$, which forms a valid MH chain on the full target, $F^H_1(\theta)$, at the root.

To ensure reversibility of the top-level chain, the $N$ scenarios can be shuffled before each step at the root-node. The subsets at each non-leaf node always contain the union of the subsets at their child nodes. Less extensive shuffles can also provide reversibility; for example when processing time series, the data can remain ordered so each `subset' is a contiguous sub-sequence; it is sufficient to randomise the order in which child sub-sequences are processed.  

\begin{algorithm}[t]
\caption{HINTS ($\text{hints\_move})$}\label{alg:HINTS}
\begin{algorithmic}[1]
\State Input: $\theta_{in}$ current state
\State Input: $h$ level in tree (0 for leaf, $\ldots,$ $H$ for root)
\State Input: $j$ node counter at this level (0 for root)
\vspace{0.25em}
\If{$h=0$} \Comment{At leaf node}
    \State $\theta'\sim q\left(\theta'|\theta_{in}\right)$ \Comment{Primitive proposal}
    \vspace{0.1em}
    \State $\Psi \gets \frac{q(\theta_{in}|\theta')}{q(\theta'|\theta_{in})}$ \Comment{Asymmetry of the proposal}
\Else  \Comment{Build a composite proposal at non-leaf}
    \State $\theta, \Psi \gets \theta_{in},1$ \Comment{Start state for the embedded chain}
    \For{$k=1,\ldots,m_h$}   \Comment{Step through child nodes}
        \State $\theta,\Psi_k\gets\text{hints\_move}(\theta,h-1,\text{child}(h,j,k))$ \Comment{Depth first traversal}
        \State $\Psi\gets\Psi\cdot\Psi_k$ \Comment{Multiply composite asymmetry}
    \EndFor
    \State $\theta' \gets \theta$ \Comment{Composite proposal}
\EndIf
\vspace{0.25em}
\State $\alpha \gets 1 \wedge \frac{F^h_j(\theta')}{F^h_j(\theta_{in})} \cdot \Psi$ \Comment{Acceptance probability}
\If{$\alpha > u \textrm{ with } u\sim\mathbb{U}[0,1]$} \Comment{Apply MH rule} 
\vspace{0.1em}
    \State $\theta_{out}, \Psi_{out}\gets\theta', \frac{F^h_j(\theta_{in})}{F^h_j(\theta')}$ \Comment{New state at this level}
\Else  \Comment{Reject}
    \State $\theta_{out},\Psi_{out}\gets\theta_{in},1$ \Comment{State is unchanged}
\EndIf
\vspace{0.25em}
\State Output: $\theta_{out}$ new state
\State Output: $\Psi_{out}$ asymmetry
\end{algorithmic}
\end{algorithm}

We found that the computational burden of HINTS could be reduced significantly by only visiting a proportion of the $m_h$ child nodes, randomly selected at each branching point. When we use this `downsampling', we always visit half of the child nodes.

Choosing parameters such as branch factor, minimum subset size, and whether to downsample, can be helped by inspecting the acceptance rates at each level in the HINTS hierarchy. To minimise computation per step a high branch factor, large leaf subsets, and downsampling are all helpful. But if a poor acceptance probability for the proposals passed up from one level to the next is obtained, this implies more work should be done to pre-screen them. Based on this, we found a configuration that works well in a wide variety of tasks: a branch factor of 4, downsampling by 2 and a leaf subset size of $N/16$. A slightly modified configuration will be specified later when we introduce proxies for HINTS.

\subsection{Subset MCMC: Summary}


We have described a number of MCMC sampling approaches that make use of subset evaluations in a variety of ways: naively subsampling, adapting the subset size using statistical tests, and building directed proposals from the subset evaluations. Many variations and improvements have been proposed for the samplers listed here. Often these are more specialised to specific likelihood structures, and most rely on smoothness assumptions. It was not possible to include all in our experimental comparison, but we review some further examples in \Cref{sec:discussion}.


\section{Subset Samplers for Irregular Likelihoods}
\label{sec:improvements}


This section first introduces two new data-driven proxies and describe how they can be integrated into each of the subset samplers (including HINTS, where this is the first example of a proxy being used). We then present a cost-aware formulation of adaptive MCMC, which takes account of the variable computational cost when selecting a proposal scale factor in subset MCMC samplers. We perform an experimental comparison against fixed scale parameters to verify that computationally-efficient sampling is achieved. We then explore the performance of the improved subset samplers on a relatively `easy' task, to verify the expected benefits, before moving on to irregular likelihoods in \Cref{sec:evaluation}. 

\subsection{Data-Driven Proxies}
\label{subsec:dd_proxy}


\subsubsection{Quadratic Proxy}

We introduce a proxy that is fitted using least-squares to states and proposals collected by the sampling chain. We aim to replace the Taylor-expansion based quadratic proxies/bounds that are used to increase efficiency in several of the subset samplers. We note that any proxy fitted separately to each scenario using \emph{least-squares} has a useful additive property: if the likelihood function is available for subsets spanning the full set of data, the coefficients of the least-squares fit for the total likelihood are identical to the sum of the coefficients of the subset fits. This means that the proxy is entirely consistent between levels in a hierarchy (for the HINTS sampler) or between partial and total evaluations (in Confidence and Austerity). This additive property does not hold for `regularised' regressions such as ridge regression. 

\begin{figure}[t]
\subfloat[Single scenario.\label{fig:SingleScenarioProxy}]{
    \includegraphics[width=1\linewidth]{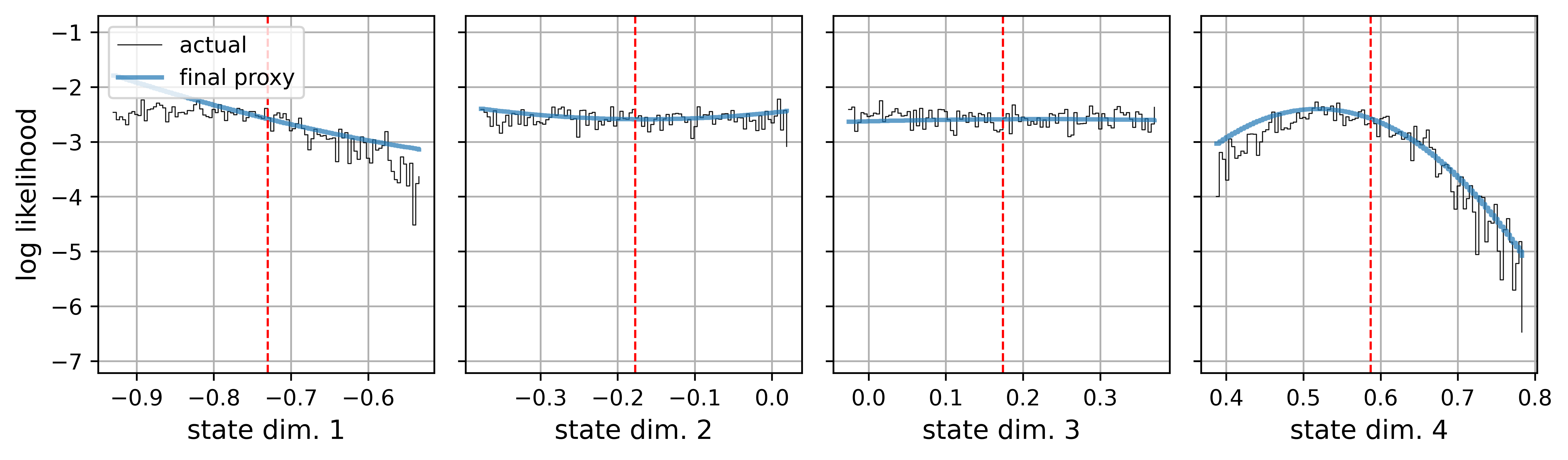}}
\\[-1ex]
\vfill
\subfloat[All scenarios combined.\label{fig:TotalProxy}]{
    \includegraphics[width=1\linewidth]{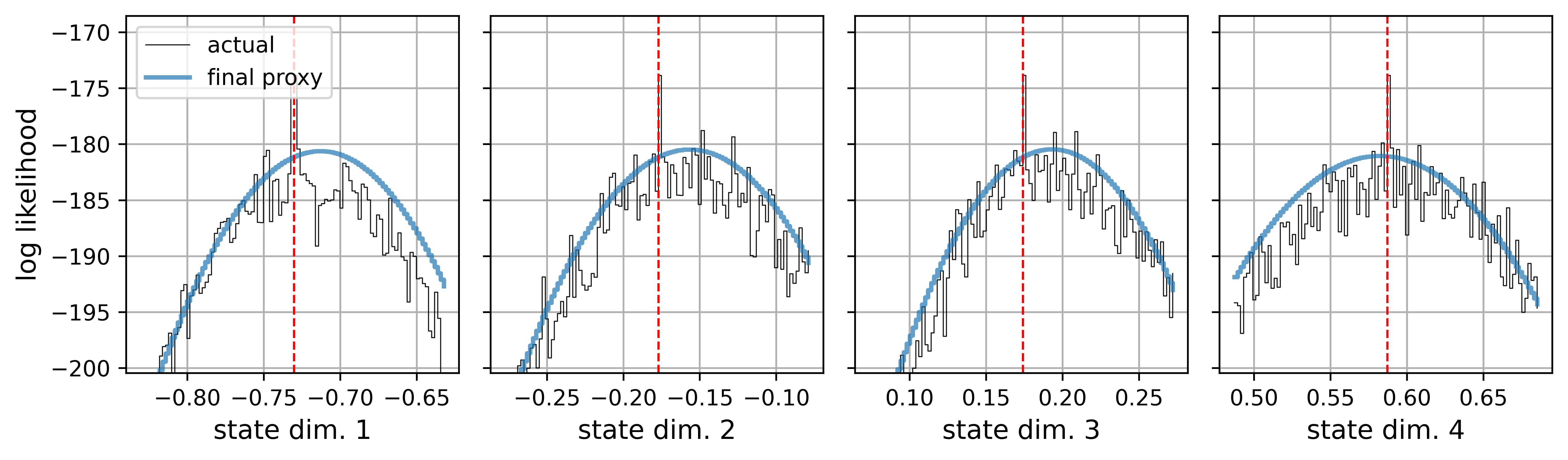}}
\caption{Quadratic proxies fitted by least-squares: conditionals in each of the 4 dimensions.}
\end{figure}

Our data-driven proxy is a global quadratic fit to not only the states output by the sampling chain so far in a chain's history, but also the proposals that were rejected. Including rejected proposals is important because proxies are used to make more reliable decisions about \textit{proposed} moves. \Cref{fig:TotalProxy} shows that a quadratic proxy fitted during a successful sampling run is reliable despite individual scenario proxies (\Cref{fig:SingleScenarioProxy}) being affected by noise. 

We always refit the proxy on full sets of scenario evaluations, for computational efficiency and simplicity, but the coefficients are stored separately for each scenario. The coefficients for the fit to any subset (or the full set) can then be easily constructed using the additive property. We assume the computations associated with fitting and using this simple proxy require arbitrarily less computation than likelihood evaluations. We do, however, account for any computation burden associated with additional likelihood evaluations undertaken for the proxy refit.

The proxy is not refitted on every time step, as this would break the reversibility of the sampling chain. Instead, it is refitted at geometrically increasing intervals (in terms of accumulated computational burden). For exact samplers (full MCMC, Firefly \& HINTS) a `strict' sampling run can be specified; this freezes the proxy (prevents refits) during the second half of the run to ensure reversibility.

\subsubsection{Nearest Neighbour Proxy}

For some tasks the quadratic proxy will fail due to non-Normal structure in the target density. We consider another generic option: a nearest neighbour (NN) regression that builds an accurate piecewise-constant representation of the explored region of the target density. This is an example of a broader class of `memory-based' (nonparametric) regression approaches that  also include kernel regression and locally-weighted regression \citep{Cizek2020Robust}. A key characteristic of our NN proxy is that it is not trusted to extrapolate outside the explored region and so proposals into unexplored states are always evaluated using the actual likelihood. Therefore it could be called a `partial proxy'. Such a proxy can act like a cache, providing a low-cost shortcut for navigating regions of the state space that are already explored.

A NN proxy with distance cut-off is able to avoid re-evaluating states that are similar to those already encountered (according to a Euclidean distance metric). We take only 1 nearest neighbour, and so this proxy replicates the target in heavily explored regions. We set the distance cut-off to a multiple (5) of the default proposal scale ($\sigma_0$) for each task. When the proxy is queried for a state beyond this distance from any previous evaluation, an evaluation of the genuine likelihood is triggered and so some computational cost is incurred.

This proxy is a very attractive option for multi-modal and other `complex' targets because it never screens-out proposed jumps into unexplored regions. It has the disadvantage that the likely achievable computational gain is lower than a well-fitted global proxy, particularly in high dimensions. Memory overhead grows linearly with the number of evaluations $T$ in the top-level chain. Computational cost of each query is $O(\log T)$ using a standard Ball Tree implementation, and remains small relative to the actual PMCMC-type likelihood evaluations.

In summary, we work with two extremes: a global parametric proxy that can make predictions across the whole state space, and a locally-trusted proxy that is much more `conservative'. In practise the best proxy for a given task may fall somewhere between these two extremes but, for research purposes, they are efficient, easy to implement, and informative for understanding the value of subset evaluations.

\subsubsection{Integration into the Subset Samplers}

For samplers that already made use of a proxy, only minor changes are needed to use the data-driven proxy in place of Taylor expansions. In each sampler run, the proxy is not used until after the first successful fit to the collected states and proposals.

For the quadratic proxy we use a full-rank quadratic fit and so the degrees of freedom scales as $O(D^2)$ for $D$-dimensional state spaces. The first proxy fit is triggered when the number of unique chain states exceeds the degrees of freedom. Subsequent refits take place at geometrically increasing intervals when expressed in terms of likelihood evaluations (computational budget consumed) rather than iterations. The geometric growth rate is 1.1. When new states and proposals generated by the chain are added to the proxy training data, an amount of data equivalent to 25\% of the new data is removed from the front of the training set; this ensures that any extremely low likelihood states encountered during `burn-in' are eventually dropped.

We use the same least-squares quadratic fit for the lower bound in Firefly, but with a constant term chosen to ensure the whole quadratic surface fits \emph{below} the likelihoods of the states being fitted. When the quadratic fits tightly under the sample distribution, Firefly is able to generate exact samples while only evaluating likelihoods for a small subset of the data. We note, the NN proxy is not able to provide a lower bound and so is not used with Firefly.

The NN proxy follows the same refitting schedule as the quadratic, except that it can be used from the start of each run and there is no need to drop burn-in evaluations. Many proposals early-on will not be within the cut-off distance of a previous evaluation and so will trigger actual likelihood evaluations. 

\subsubsection{Integration of Proxies into HINTS}
HINTS was proposed without a proxy or control variate. We add data-driven proxies (evaluated on subsets of scenarios) into HINTS to achieve more efficient mixing. Starting with a (log) subset information function $\log F^h_j(\theta)$ that contains a sum of scenario log likelihoods, $\Sigma_{i \in \mathbb{X}_j^h} l_i(\theta)$, we consider four options for including proxy evaluations $\hat{l}_i(\theta)$:
\begin{align}
    \log F^h_j(\theta) &= \Sigma_{i \in \mathbb{X}_j^h} l_i(\theta) + \Sigma_{i \notin \mathbb{X}_j^h}\hat{l}_i(\theta)\label{eq:p1},\\
    \log F^h_j(\theta) &= \Sigma_{i \in \mathbb{X}_j^h} l_i(\theta) + \Sigma_{i \in \tilde{\mathbb{X}}_j^h \setminus {\mathbb{X}_j^h} }\hat{l}_i(\theta)\label{eq:p2},\\
    \log F^h_j(\theta) &= \Sigma_{i \in \mathbb{X}_j^h}\hat{l}_i(\theta)\label{eq:p3},\\
    \log F^h_j(\theta) &= \Sigma_{i \in \tilde{\mathbb{X}}_j^h} \hat{l}(\theta)\label{eq:p4},
\end{align}
except at the root where we retain $\log F^H_1(\theta) = \Sigma_{i=1}^N l_i(\theta)$.

In \Cref{eq:p1} we extend each subset likelihood evaluation with proxy evaluations for all scenarios that are not in the subset. This is the same approach used by the Confidence sampler. \Cref{eq:p2} limits the proxy evaluations to scenarios that are in the parent subset, denoted $\tilde{\mathbb{X}}_j^h$. This retains an element of tempering,\footnote{Tempering, discussed further in \Cref{subsec:tempering}, uses a distribution with lower likelihood variation to provide opportunities for large moves, before evaluating on a full distribution with larger likelihood variation.} providing robustness to an imperfect proxy.

\Cref{eq:p3} completely replaces likelihood evaluations with proxy evaluations below the root. This means computation per iteration is reduced to essentially the same as MCMC with the global quadratic proxy, despite constructing much larger directed proposals. Finally, \Cref{eq:p4} expands these proxy-only evaluations to the scenarios that are present in the immediate parent subset. For example, with 64 scenarios and a branch factor of 4, the root-node proposals will be the composite of 4 steps accepted or rejected with the full proxy. Each of these 4 steps are themselves using proposals built from up 4 steps accepted or rejected with the proxy on 16-scenario subsets.  

While all of these options were tried, we found no evidence that (costly) evaluations of the actual likelihood are useful below the root level once the proxy is established. \Cref{eq:p4}, which pre-screens proposed moves with the proxy while retaining the natural tempering structure, was most effective for the quadratic proxy; this is the configuration we evaluate fully. The NN proxy only provides `cheap' evaluations in regions that have been explored, otherwise it falls back to an actual likelihood evaluation. This can make proxy evaluation of the full parent set expensive and so \Cref{eq:p3} was chosen. 

In summary, once a proxy has been established, the function evaluations on line 13 of \Cref{alg:HINTS} are replaced with proxy evaluations, except at the root-node. In our experiments this means the smallest batch of proxy evaluations is $N/4$ (quadratic) or $N/16$ (NN). Downsampling was not used with the quadratic proxy because there is no significant overhead in additional proxy evaluations and visiting all scenarios makes composite proposals more reliable.

\subsection{Computation Cost-Aware Adaptive MCMC for Subset Samplers}
\label{subsec:computation_adaptive}

Existing adaptive MCMC approaches (\Cref{subsec:adaptive_MCMC}) have several drawbacks in the context of subset samplers and irregular likelihoods:

\begin{enumerate}[itemsep=1pt]
\item Samplers that make use of subset evaluations may use less computation per iteration with some step sizes than with others. In order to obtain the best sample within a given computational budget, it is desirable for the adaptation algorithm to take into account the computation cost (expected number of likelihood evaluations) for each possible choice of the random walk proposal parameter $\sigma$.
\item The optimal acceptance probability is not consistent between tasks or between samplers on a given task (confirmed in \Cref{subsubsec:validation}). It is also dependent on the level of correlation between state dimensions. This means we would either have to repeat experiments to find the best acceptance rate target, or risk an unfair comparison between samplers.
\item With irregular likelihoods that have noise at specific spatial scales, the assumed monotonic decreasing relationship between acceptance probability and step size can break down at small scales. There is then a risk that an adaptive algorithm targeting a fixed acceptance probability will diverge towards an extremely small step size. This leads to arbitrarily long mixing times and the failure to obtain an accurate sample. This was easily verified with our experimental tasks.
\end{enumerate}

\subsubsection{Squared Jump Distance per Unit of Computation}

We aim to find an adaptive MCMC approach that can be operated with the same fixed parameters regardless of the task or sampler. Methods based on Expected Squared Jump Distance (ESJD) have been shown to be more robust than fixing a target acceptance probability \citep{pasarica2010adaptively}. ESJD is the empirical average of the squared Euclidean distance between successive states output by the sampling chain, including zeros for rejections. Being closely related to autocorrelation, ESJD is a good predictor of mixing performance. Maximising ESJD may lead to, for example, operation with a large step size and low acceptance rate (or vice versa) on a \textit{task-specific} or \textit{sampler-specific} basis.

With full knowledge of the proposal density, it is possible to predict the impact of a change to the proposal scale parameter using importance sampling (IS). This works by re-weighting the observed ESJD values according to the relative probability of selecting each proposal under a different (scale) parameter, compared with the one used at the time. The IS mechanism is not appropriate in all contexts because the importance weights may not be readily available or reliable for samplers that build composite proposals (\textit{e.g.,} HINTS) or for more specialised proposal mechanisms.

We propose that adaptive control should take into account computational cost trade-offs for samplers that use subset evaluations. Samplers like Austerity use statistical tests on subset likelihood evaluations; they only evaluate on a larger set if necessary and this means large proposed moves may require fewer computations on average to accept or reject them. The dependency of computation on proposal scale is expected to be task-dependent and sampler-dependent. In HINTS, for example, computation burden is variable depending on rejection rates at each level in the hierarchy (due to re-use of identical evaluations or use of proxy evaluations). Therefore, our adaptive control algorithm has the objective of maximising ESJD \textit{per likelihood evaluation}. This can be interpreted as the variance rate (or volatility) of the state \emph{per unit computation} and we will see it is a good predictor of mixing performance. 

In all the samplers we consider, the acceptance probability is known for each top-level proposal. Therefore, a reduced-variance ESJD is obtained using this probability as a weight for the proposed jump, rather than the $\{0,1\}$ outcome of the acceptance decision.

\begin{algorithm}
\caption{Computation-cost aware adaptive MCMC: choose action at step $\tau$}\label{alg:adaptive}
\begin{algorithmic}

\State History: $d_{0:\tau-1} = \left(\|\theta_t - \theta'_t\|^2\right)_{t=0}^{\tau-1}$ \Comment{squared proposed jump distance at each step}
\State History: $k_{0:\tau-1}$ \Comment{chosen action index at each step}
\State History: $\alpha_{0:\tau-1}$ \Comment{acceptance probability at each step}
\State History: $c_{0:\tau-1}$ \Comment{computational resource consumed at each step}
\vskip 3pt
\State $t_1 \gets \left \lfloor{\tau/4}\right \rfloor$ \Comment{start of time range for analysis}
\For{$k=1,\ldots,K$} \Comment{gather statistics for each possible action, $k$}
    \State $N_k \gets \sum_{t = t_1}^{\tau-1}{1\!\!1}_k(k_t)$ \Comment{number of times action $k$ has been taken}
    \State $D_k \gets \sum_{t = t_1}^{\tau-1}{1\!\!1}_k(k_t) \alpha_t d_t$ \Comment{expected total jump distance for $k$}
    \State $C_k \gets \sum_{t = t_1}^{\tau-1}{1\!\!1}_k(k_t)c_t$ \Comment{total computation for $k$}   
    \State $p_k \gets \frac{1}{N_k}\sum_{t = t_1}^{\tau-1}{1\!\!1}_k(k_t)\alpha_t$ \Comment{average acceptance probability for $k$} \vskip 3pt     
\EndFor    
\If{$\exists{k}$ such that $N_k = 0$} \Comment if any action is untaken in recent history
\State Output: $k_\tau$ uniform from set $\{k \mid N_k=0\}$ \Comment{choose random untaken action} \vskip 2pt
\ElsIf {$\epsilon > u  \textrm{ with } u\sim\mathbb{U}[0,1]$} \Comment{exploratory action w.p. $\epsilon$}
\State Output: $k_\tau$ s.t. $P(k_\tau = k) \propto \frac{N_k + 1}{C_k}$ \Comment{choice inversely proportional to cost}
\Else \Comment{greedy action subject to acceptance constraint}
\State Output: $k_\tau \gets \underset{\{k \mid p_k > \alpha_0\}}{\mathrm{argmax}} \frac{D_k}{C_k}$ \Comment{Maximise ESJD per likelihood evaluation}
\EndIf
\end{algorithmic}
\end{algorithm}

\subsubsection{Proposed $\epsilon$-Greedy Algorithm}

To avoid importance sampling, we only consider a fixed set of multipliers $\{r_k\}_{k=1}^K$ for $\sigma_0$. Choosing between a fixed set of actions is a \textit{bandit problem} familiar in reinforcement learning \citep{sutton_barto_1998}. During each sampling run we accumulate, for each action, the total squared jump distance, and the total computational cost expended. The ratio of these quantities is the ESJD per likelihood evaluation (the objective to maximise). We use an $\epsilon$-greedy policy \citep{watkins1989learning}; this selects the best action on this objective with probability $(1 - \epsilon)$ and an `exploratory' action with probability $\epsilon$. 

The exploratory actions are selected randomly, but with probabilities that are proportional to the inverse of the expected computational cost; this avoids actions with high computational costs (\textit{i.e.,} those that tend to use more likelihood evaluations) from consuming more than their fair share of the exploratory computation budget. Untaken actions are assumed to have zero computational cost and so are prioritised.

For all the samplers, a new choice of $r$ is drawn once for each step of the top-level chain, even if (as in HINTS) multiple proposals are used between top-level decisions. \Cref{alg:adaptive} details how each such action is chosen. It makes use of quantities collected over the sampler history; relevant data for each action is selected using the operator ${1\!\!1}_k(k_t)$ which returns 1 when $k_t = k$ and 0 otherwise. 

In some samplers, rejections made with only a proxy do not require likelihood evaluations, and so are regarded as consuming no computational resource, but there will be some overhead in running chains for an unbounded number of iterations. Therefore, we do not select greedy actions with expected acceptance probability of less than 2\% ($\alpha_0 = 0.02$). Where computations are performed that are not directly attributable to an action (\textit{e.g.,} Firefly auxiliary resampling), the running mean of this `overhead' is added to every action.

For the comparison in \Cref{subsubsec:validation}, four actions represent $r \in \{0.5, 1, 2, 5\}$, and each of these multipliers is also evaluated separately in a fixed-scale run. In our main experiments, we use $K=11$ values for $r$ in $[0.1,10]$ (inclusive) equally spaced on a $\log$ scale. Throughout, we use $\epsilon = 0.1$ and ignore the first 25\% of the (running) history, so as to progressively drop burn-in moves. 



Integrating this adaptive algorithm into most of the samplers was straightforward: a counter is used to keep track of the number of unique likelihood evaluations used to build the proposal (where applicable) and to make the accept or reject decision.

For HINTS, many `primitive' proposals may be made inbetween root-node steps, but a new action (scale factor choice) was only made once for each such step. Furthermore, when operating with a proxy we may see `zero moves' of the top-level chain where the composite proposal from the lower levels returns the input state, which always gets accepted. These do not impact the action evaluation because they contribute zero squared jump distance and zero evaluation cost, but we treat them as rejections for the purpose of enforcing the lower bound on acceptance probability.

\subsubsection{Experimental Validation}
\label{subsubsec:validation}


We performed a large comparison of the proposed cost-sensitive adaptive step size algorithm against four fixed proposal scales for all samplers, running with and without proxies, where appropriate. The task chosen was the noise-free version of a 4D synthetic test task (\Cref{sec:evaluation} and \Cref{app:synthetic}); this was intended to be sufficiently `easy' that all of the subset samplers could achieve a speed-up over MCMC. The quadratic proxy is suitable for this task.

\begin{table}[t!]
\centering
\addtolength{\tabcolsep}{-0.2em}
\begin{tabular}{l|rrrrr|rrrrr|rrrrr}
\toprule
\fbox{4D noise-free} & \multicolumn{5}{c}{accept \%} & \multicolumn{5}{c}{evals/step} & \multicolumn{5}{c}{variance/eval} \\
\midrule[0.05pt]
multiplier & 0.5 & 1 & 2 & 5 & A & 0.5 & 1 & 2 & 5 & A & 0.5 & 1 & 2 & 5 & A \\
\midrule
MCMC & 72 & 48 & 20 & 2 & 22 & 1.00 & 1.00 & 1.00 & 1.00 & 1.00 & 0.7 & 1.6 & 1.9 & 0.4 & 1.8 \\
Firefly & 72 & 49 & 20 & 2 & 22 & 0.18 & 0.17 & 0.16 & 0.20 & 0.17 & 3.7 & 9.0 & 11.9 & 2.2 & 10.4 \\
HINTS & 58 & 35 & 16 & 2 & 33 & 1.63 & 1.50 & 1.25 & 0.74 & 1.46 & 0.9 & 1.4 & 1.4 & 0.8 & 1.4 \\
HINTS+proxy & 96 & 82 & 57 & 15 & 17 & 0.97 & 0.83 & 0.57 & 0.16 & 0.18 & 5.5 & 9.4 & 13.1 & 19.4 & 17.6 \\
Austerity & 70 & 50 & 27 & 6 & 11 & 0.79 & 0.79 & 0.71 & 0.49 & 0.54 & 0.8 & 2.1 & 4.2 & 5.6 & 4.4 \\
Confidence & 72 & 49 & 21 & 2 & 22 & 1.00 & 1.00 & 1.00 & 1.00 & 1.00 & 0.7 & 1.6 & 1.9 & 0.5 & 1.8 \\
Conf+proxy & 72 & 49 & 20 & 2 & 22 & 0.17 & 0.17 & 0.14 & 0.08 & 0.14 & 3.8 & 9.2 & 13.3 & 5.6 & 12.8 \\
\bottomrule
\end{tabular}
\caption{Comparison of mixing performance between fixed values (0.5, 1, 2, 5) and adaptive control (A) of proposal scale multiplier, $r$.}
\label{tab:adaptive}
\addtolength{\tabcolsep}{0.2em}
\end{table}

The metrics are detailed in \Cref{app:performance_metrics}; we report the median of 50 parallel runs, with a measurement interval that is the second half of each run (by computational resource consumed). In the table, `variance/eval' shows the total realised squared jump distance per full data likelihood evaluation. We will refer to this in the text as \textit{variance rate}.

The results in \Cref{tab:adaptive} show that the adaptive algorithm can achieve a variance rate that is competitive with the best (in hindsight) fixed step size, but with an `overhead' cost associated with selecting sub-optimal actions early in the run and continuing to use 10\% exploratory actions. For example, the best sampler achieves a variance rate of 17.6 (adaptive) which is 91\% of 19.4 achieved by the best fixed multiplier, $r=5$. Asymptotically the `greedy' action will be the best, and so we expect the exploration overhead to be no more than 10\% for long runs; the results appear consistent with this. The computational gain compared with exhaustive evaluation of all $K$ multipliers is therefore approximately $K(1-\epsilon)$. Other metrics used to monitor mixing performance (\textit{e.g.,} effective sample size per evaluation) showed a similar pattern. We avoided inspecting our final sampling accuracy metric at this stage. 

Depending on the sampler, the adaptation can lead to running at a large proposed step size with a smaller acceptance rate, or vice versa. For example, `HINTS+proxy' runs with a very large step size and low acceptance rate (17\%) on this task\footnote{The same acceptance rate could have been measured at 99\% using a different convention (ignoring zero-cost moves), and so in general, acceptance probability is not a well-defined objective. Therefore, in \Cref{sec:evaluation} we will combine the first 2 metrics to report the number of acceptances \textit{per full evaluation}.}. As a result it makes fewer computations per step, because rejections that are made using a proxy do not require likelihood evaluations. Any adaptive algorithm that targets a fixed acceptance rate, or more broadly any algorithm that is not computation-aware, would be unlikely to find this solution. We also see a preference for a lower acceptance rate (larger multiplier) in the Austerity sampler, for a different reason: less computation is needed to evaluate large likelihood changes in approximate samplers that use statistical tests. 

Note in this 4D task that, once the scale factor has been optimised in each case, the Confidence sampler \emph{without} proxy behaves in the same way as full MCMC: at the optimised proposal scale it is not able to make decisions with less than a full sweep of the data. Therefore, in subsequent experiments, we will always use a proxy with the Confidence sampler. In contrast, the Austerity sampler is able to achieve a high variance rate at a high acceptance rate, by using a less `conservative' statistical test.

Although we compared our adaptive algorithm with non-adaptive (fixed $\sigma$) runs, we undertook some runs with an adaptive controller that targets a specific acceptance rate, and verified that performance was similar to the fixed multiplier yielding the same acceptance rate. This is not surprising because once burn-in is complete, and the `learning rate' for adaptation has reduced, the two are equivalent.

The adaptation mechanism could be improved for specific samplers, but is robust and serves the purpose, for this research, of providing a \textit{fair comparison} of a range of samplers. 

\begin{figure}[p]
    \subfloat[Austerity has fast burn-in but then needs larger subsets 
    as likelihood changes become smaller.]{\includegraphics[trim={0 2ex 0 0},clip,width=0.48\linewidth]{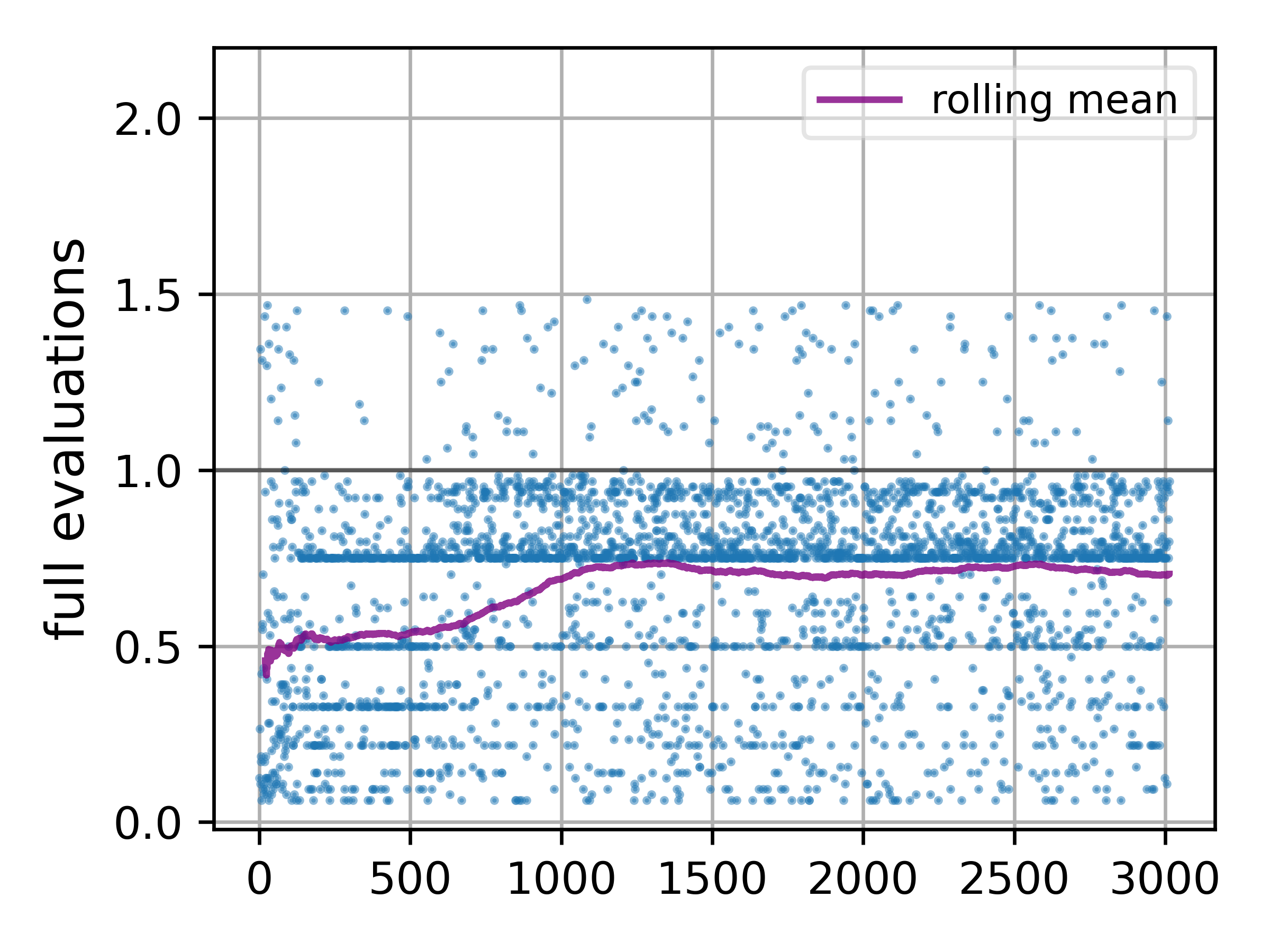}\label{fig:AusterityEvals}}
    \hfill
    \subfloat[Confidence is very efficient once an accurate proxy is established at $\sim 1000$ iterations.]{\includegraphics[trim={0 2ex 0 0},clip,width=0.48\linewidth]{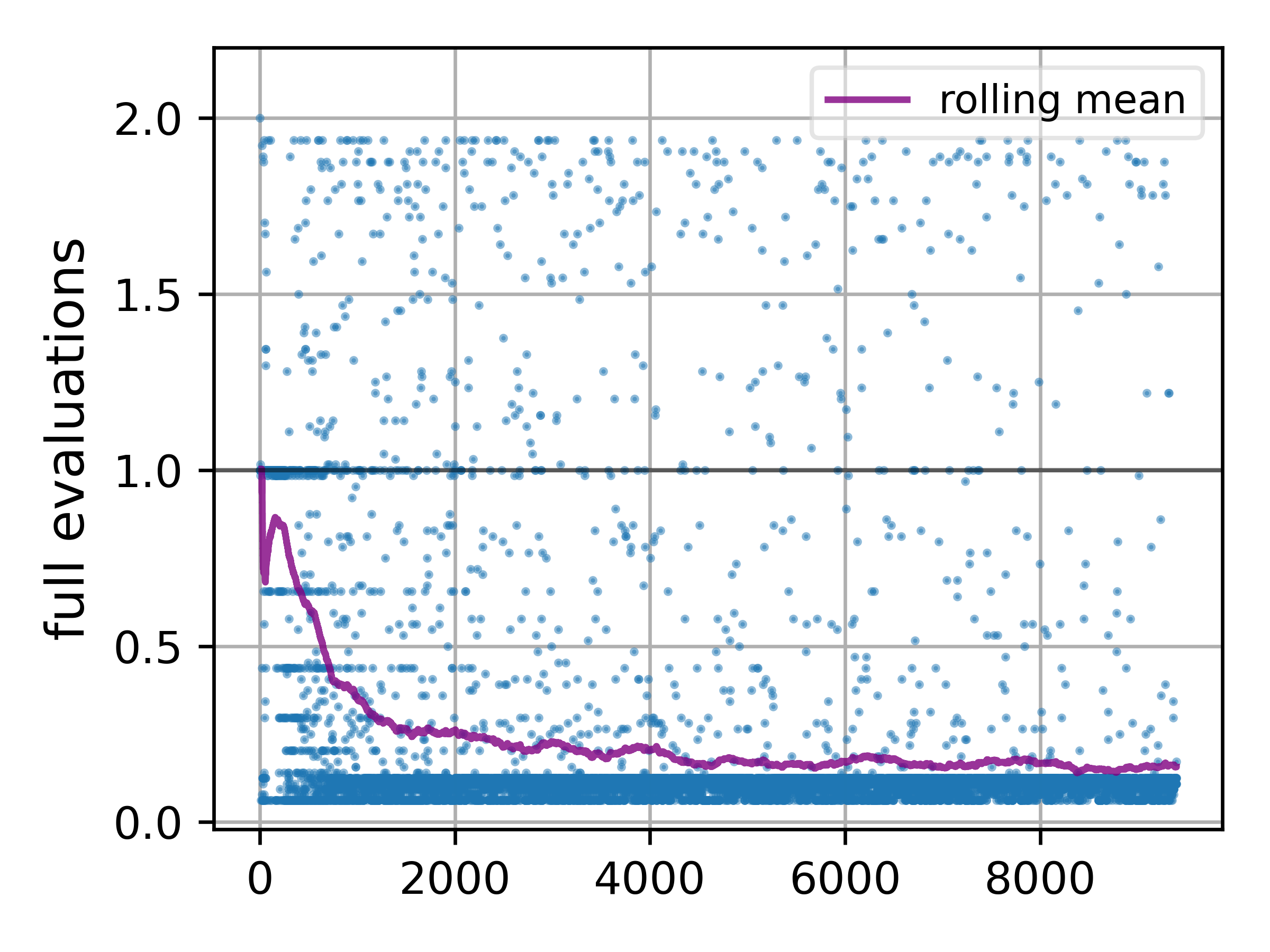}\label{fig:ConfidenceEvals}} 
    \\[-1ex]
    \subfloat[Firefly evaluates progressively fewer likelihoods as tight lower bounds are gradually established.]{\includegraphics[trim={0 2ex 0 0},clip,width=0.48\linewidth]{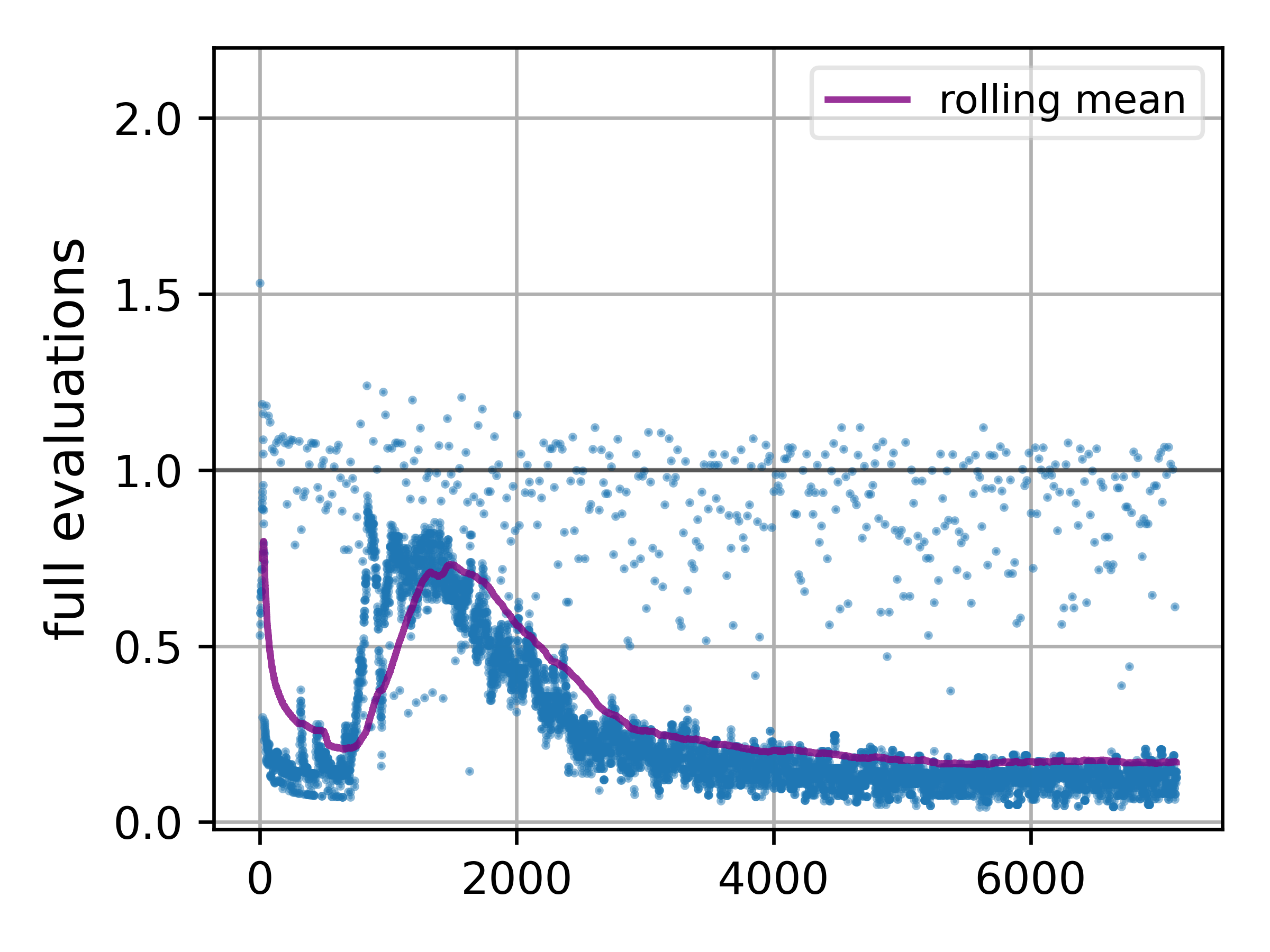}\label{fig:FireflyEvals}}
    \hfill
    \subfloat[HINTS \textit{without} proxy uses $>$1 full evaluation to construct then accept/reject each directed proposal.]{\includegraphics[trim={0 2ex 0 0},clip,width=0.48\linewidth]{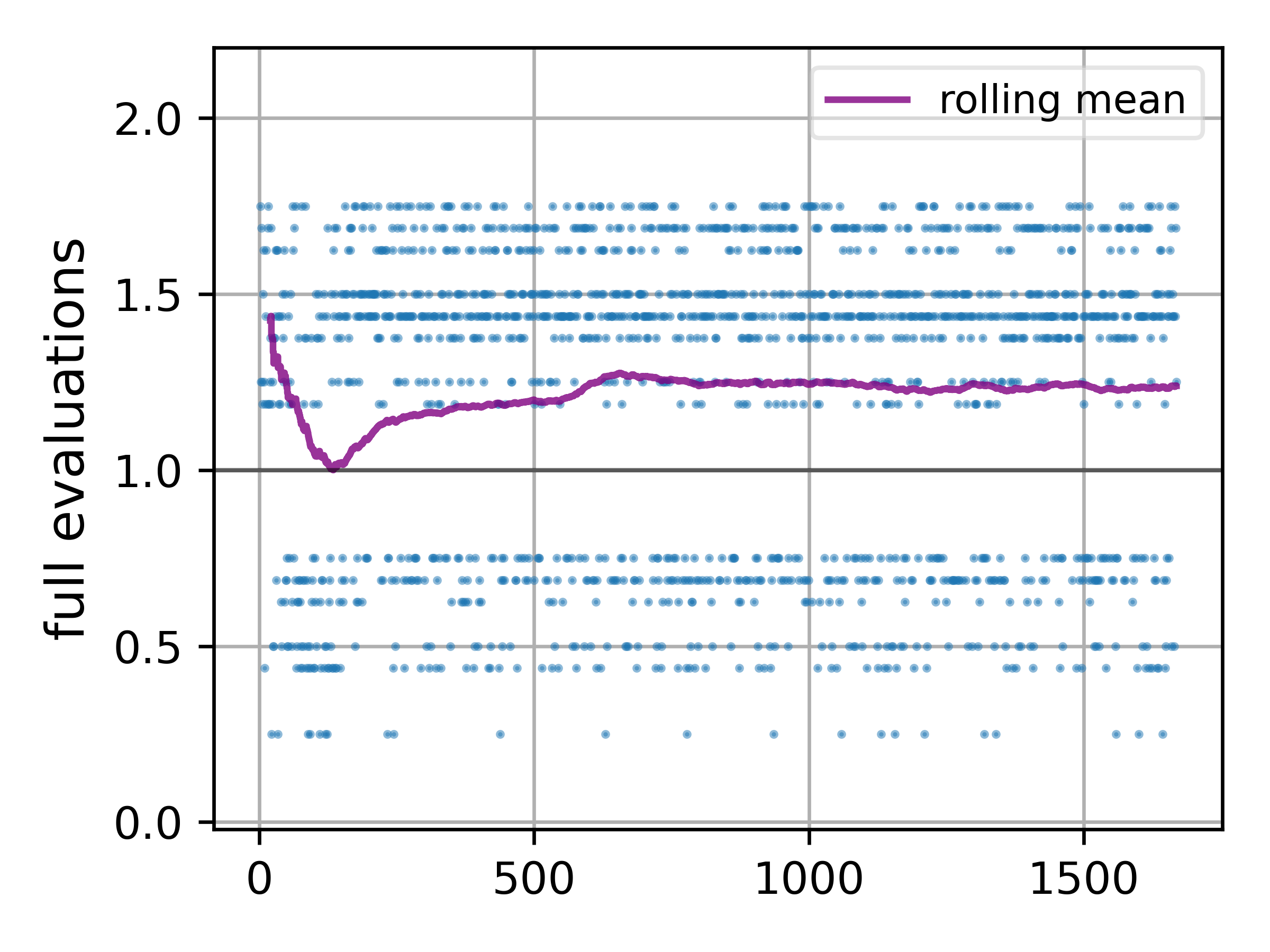}\label{fig:HINTSEvals}}
    \\[-1ex]
    \vfill
    \subfloat[\textit{Aside}: HINTS \emph{with proxy} chosen actions (proposal scale factors); first 1000 iterations shown.]{
    \raisebox{3mm}{
    \dbox{\includegraphics[trim={1ex 2ex 0 1ex},clip,width=0.44\linewidth]{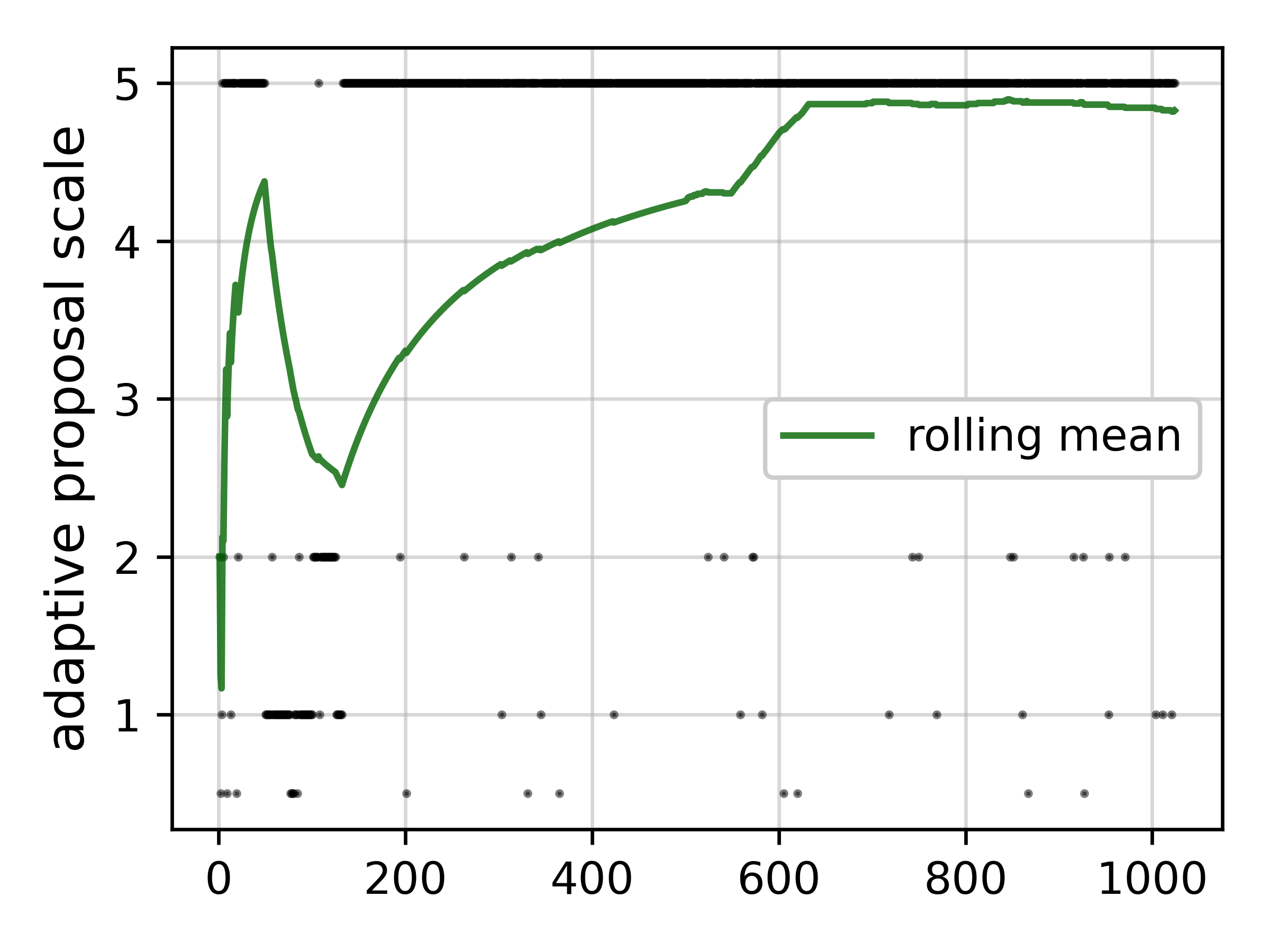}\label{fig:HINTSproxyadapt}}}}
    \hfill
    \subfloat[HINTS \emph{with proxy} only uses actual evaluations for proposals not screened out by the proxy.]
    {
    \includegraphics[trim={0 2ex 0 0},clip,width=0.48\linewidth]{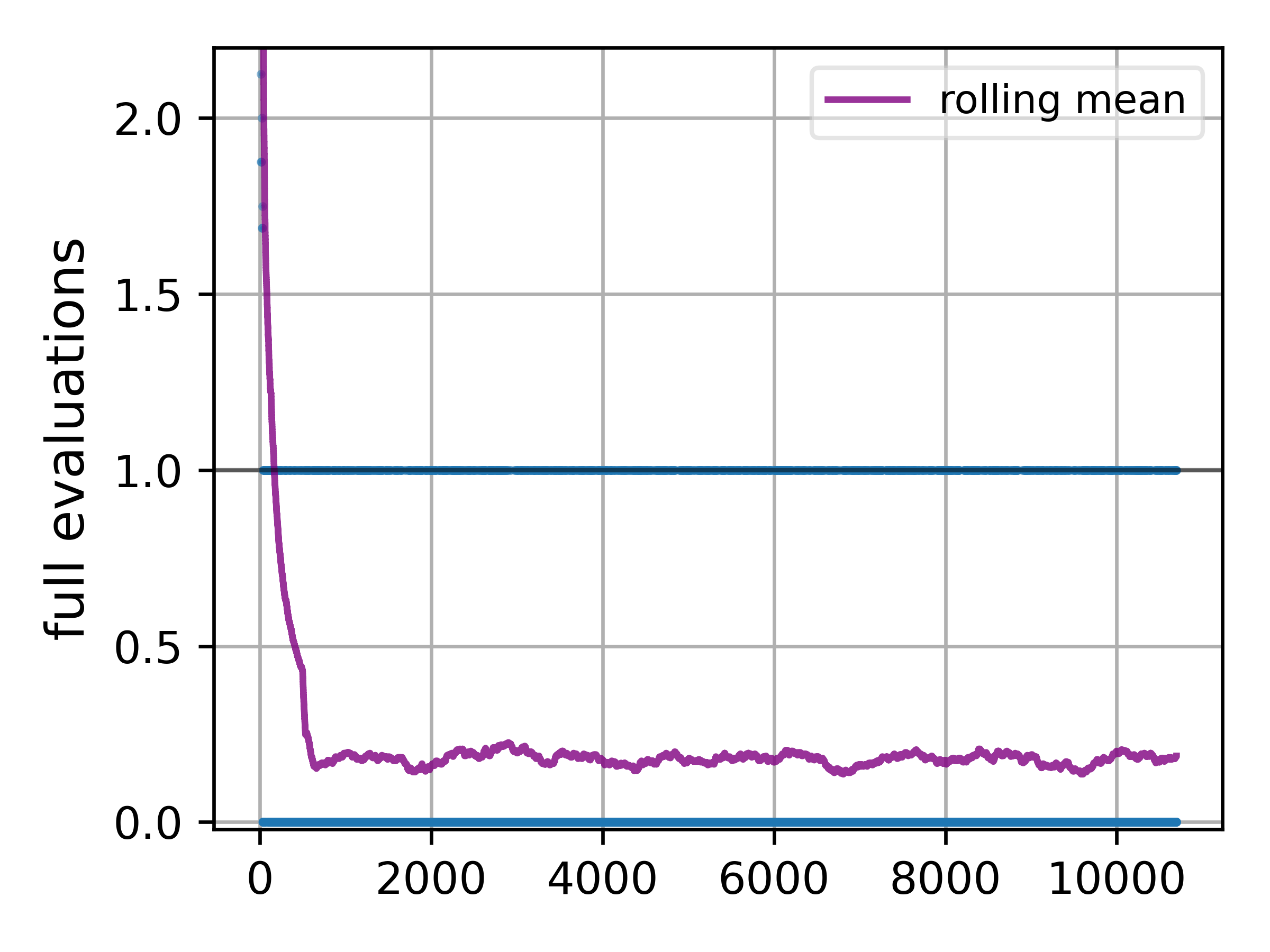}\label{fig:HINTSproxyEvals}}
    \caption{(a)--(d): proportions of the data set evaluated on each step by existing samplers; full MCMC = 1. Each uses a different number of iterations (x axis) for an equivalent computation budget. Integration of a proxy enables HINTS to use (e) a large proposal scale \emph{and} (f) much less computation per iteration.}
\end{figure}



\subsection {Testing the Improved Samplers} 
\label{subsec:modified_sampler_testing}

\Cref{sec:evaluation} will compare sampling accuracy on a broader range of tasks but here we explore the ability of the improved subset samplers to obtain a computational advantage relative to full MCMC on the `easy' task, with a data-driven proxy where applicable.


Subsample MCMC has a fixed computational gain over full MCMC, but in the more advanced subset methods, we see variable benefit during sampling. \Cref{fig:AusterityEvals} shows how Austerity usually accepts/rejects proposals with fewer full likelihood evaluations per iteration (0.61 on average) than MCMC (1.0).  Occasionally it can use more than MCMC; this happens when a state that was accepted using a small subset has to be evaluated along with a new proposal, on the full set. Once each sampling chain reaches the high-likelihood region, mixing performance is reasonable (in terms of variance rate) but the smaller likelihood changes that occur in this region mean larger subsets are needed for significance in the t-test. 

We saw in \Cref{tab:adaptive} that, without a proxy, mixing for the Confidence sampler was far worse than the Austerity sampler due to its stricter statistical test, but this reversed when the proxy was included. \Cref{fig:ConfidenceEvals} confirms that the Confidence sampler uses only 16\% of the data set per iteration on average; much less than the Austerity sampler's 61\%.









\begin{figure}[h]
    \centering
    \includegraphics[width=1\linewidth]{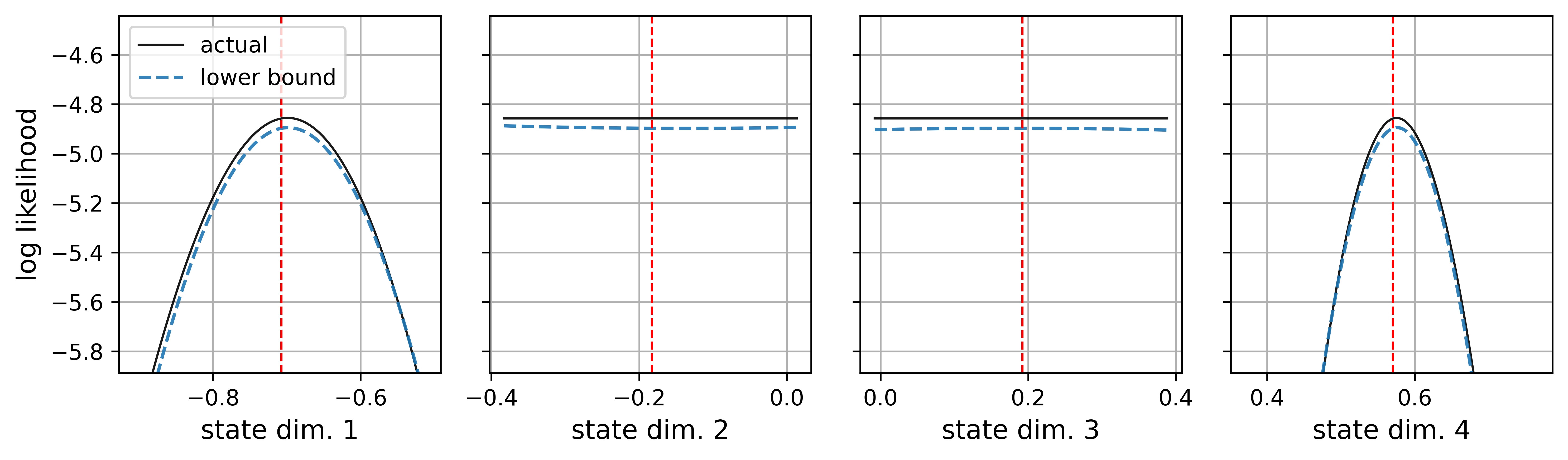}
    \caption{Firefly quadratic lower bound for single scenario, 4D noise-free task. The bounds are tight when the smooth likelihood has near-quadratic shape.}
    \label{fig:FireflyBound}
    \centering
    \includegraphics[width=1\linewidth]{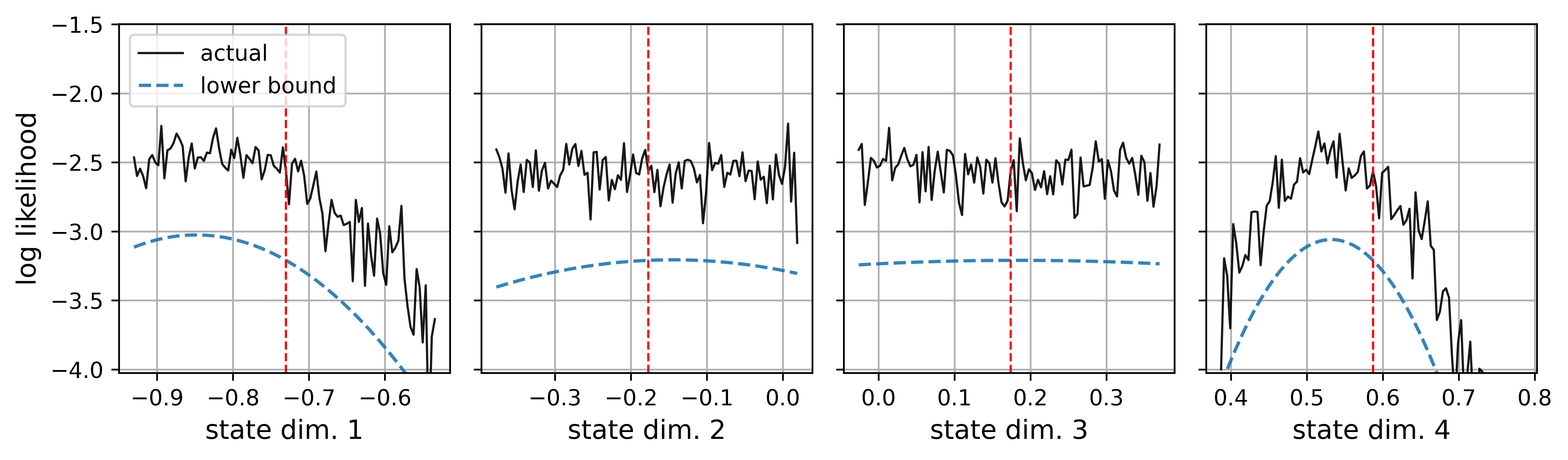}
    \caption{When there is likelihood noise, the Firefly lower bounds are not tight and so fewer scenarios will be `dark'. (The quadratic bound touches the actual likelihood at locations that do not fall in these 1D conditionals.)}
    \label{fig:FireflyBound2}
\end{figure}

Firefly, using the data-driven quadratic as a lower bound, as shown in \Cref{fig:FireflyBound}, was able to achieve a similar acceptance rate as MCMC but using 28\% as many likelihood evaluations per step, \emph{including} the overhead of making regular full evaluations to train the data-driven proxy. However, \Cref{fig:FireflyEvals} shows that it took some time to establish bounds tight enough to allow a large proportion of the scenarios to be `dark' in the mixing stage, even on this noise-free task. Later we will see that the Firefly quadratic data-driven lower bound is much less tight when there is high-frequency noise in the likelihood (see \Cref{fig:FireflyBound2}). 

\Cref{fig:HINTSEvals} illustrates the main issue with the original HINTS algorithm: multiple subset evaluations are used at lower levels of the sampling hierarchy, and despite reusing evaluations, HINTS uses more per iteration than full MCMC. In contrast, HINTS with a global quadratic proxy (\Cref{fig:HINTSproxyEvals}) uses much \emph{less} computation per iteration than full MCMC---only 18\% for the comparison in \Cref{tab:adaptive}---because after calibrating the proxy, the only circumstance in which it makes non-proxy likelihood evaluations is in the top-level chain when a non-zero composite proposal is constructed by the proxy-only hierarchical processing.

\section{Experimental Comparison of Samplers for Irregular Likelihoods}
\label{sec:evaluation}

This section evaluates the improved subset samplers on tasks with irregular likelihoods. A key motivation is Bayesian inference of the parameters of a disease model, detailed in \Cref{app:time_series_PF_example}, using a PF (\Cref{app:particle_filter}) to obtain the likelihood. We are only able to use a limited (precomputed 2D) version of the disease model, because our 50-run statistical comparison of multiple samplers on full PMCMC would require huge computational resource. Therefore we explore tasks with more dimensions, extra correlation, and tall data, using a synthetic likelihood function that mimics the PF noise characteristics. 


\subsection{Overview of evaluation tasks}\label{subsec:eval_tasks}

\begin{figure}
    \centering
    \includegraphics[trim={0 1ex 0 0},clip,width=0.75\linewidth]{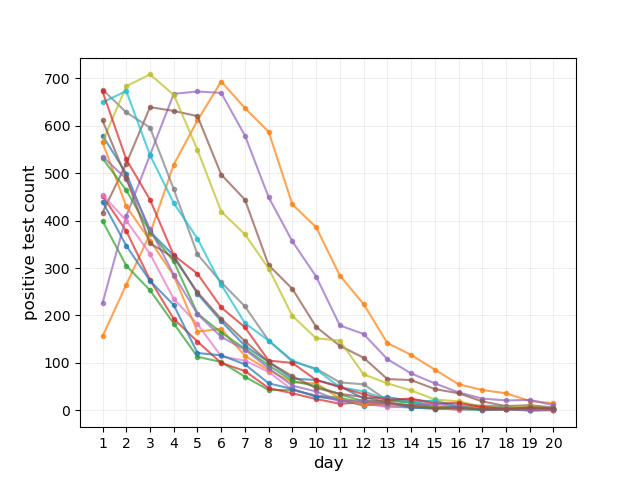}
    \caption{Sample of 16 sequences from 2D disease model.}
    \label{fig:DiseaseDataSample}
\end{figure}

Here we use a simplified form of the disease model where the E \& I and S \& R compartments have been combined. As a result, the model includes only two parameters representing the transition probabilities between these combined compartments. Some example sequences are shown in \Cref{fig:DiseaseDataSample}. Each 4000-particle run calculates the likelihood of a time series of 20 observations given model parameters $\theta$. By limiting to 2D, we have been able to completely characterise this likelihood on a precomputed grid (to 3 decimal place precision) so that many parallel runs of multiple configurations can be undertaken without actually exercising the PF for states that have been seen before. 

The `Synthetic' task detailed in \Cref{app:synthetic} has similar noise characteristics to the disease model but is fast to compute for research purposes. We still recognise that, in genuine applications, the likelihood evaluation would be expensive and so this remains our measure of computation cost. It is straightforward to vary parameters such as number of state dimensions and noise characteristics. Specifically, we implemented a zero-noise version of this task (for which all samplers are expected to be successful) for the examples in \Cref{sec:improvements}. We also implemented a version with higher correlation between state dimensions to explore the limitations of random walk proposals and non-Normal target structures. We present results for this flexible task before returning to the disease model in \Cref{subsec:disease_results}.

\subsection{Comparison of Exact Samplers}

\begin{figure}
    \centering
    \includegraphics[width=1\linewidth]{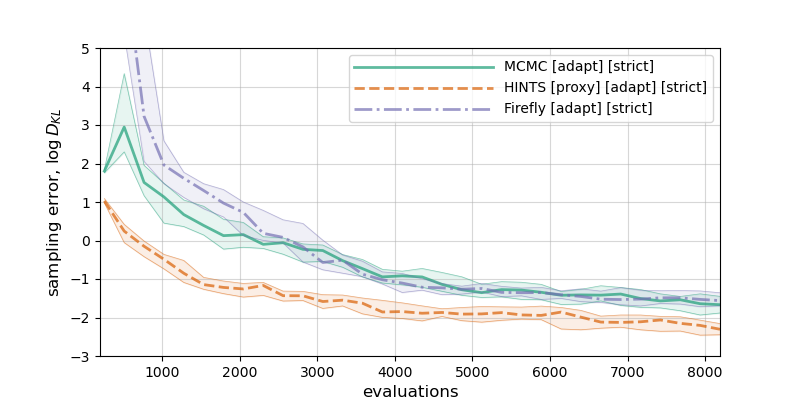}
    \caption{Comparison of sampling errors between the exact samplers on a 4D task.}
    \label{fig:Exact4DError}
\end{figure}

 We take the 4D task (with noise) as our central example because it is sufficiently challenging to separate the samplers. Exact samplers should generate true samples from the target density when run for a sufficient number of iterations. \Cref{fig:Exact4DError} shows the sampling error against computation done for selected configurations. Firefly (using the quadratic lower bound) does not show any benefit over full MCMC whereas HINTS appears to burn-in faster and continue to improve the accuracy of the sample.

We can understand this performance using additional metrics reported in \Cref{tab:Exact4D}. Firefly and HINTS without proxy perform no better than full MCMC. Firefly clearly does not benefit significantly from the lower bound because it has no more acceptances per likelihood evaluation than full MCMC. This is because the lower bound is affected by noise as shown in \Cref{fig:FireflyBound2}. The quadratic function must not be above the actual likelihood surface at any point in the 4D space, making it very sensitive to the worst-case noise.

HINTS without a proxy only achieves similar performance to MCMC on the mixing measures (variance/eval and ESS/eval), indicating that the sampling hierarchy does not add value in this case. In contrast, HINTS with a quadratic proxy achieves a significantly higher variance rate (per unit of computation) by exploiting proxy evaluations in its hierarchy to construct larger directional moves \emph{and} higher acceptance rates. Its asymptotic error (0.10) is approximately half that of full MCMC (0.20). The convergence indicator $\hat{R}$ confirms consistency between the 50 parallel runs.

\begin{table}
\centering
\begin{tabular}{lrrrrr}
\toprule
\fbox{4D} & \multicolumn{1}{c}{accept/eval} & \multicolumn{1}{c}{variance/eval} & \multicolumn{1}{c}{ESS/eval} & \multicolumn{1}{c}{$D_{KL}$} & \multicolumn{1}{c}{$\hat{R}$} \\
\midrule
MCMC & \texttt{0.07\small{ ±0.01}} & \texttt{0.85\small{ ±0.06}} & \texttt{.008\small{ ±.002}} & \texttt{0.20\small{ ±0.04}} & \texttt{1.03} \\
Firefly & \texttt{0.07\small{ ±0.01}} & \texttt{0.75\small{ ±0.07}} & \texttt{.008\small{ ±.001}} & \texttt{0.22\small{ ±0.04}} & \texttt{1.03} \\
HINTS & \texttt{0.05\small{ ±0.01}} & \texttt{0.72\small{ ±0.08}} & \texttt{.007\small{ ±.001}} & \texttt{0.22\small{ ±0.04}} & \texttt{1.03} \\
HINTS+proxy & \texttt{0.13\small{ ±0.01}} & \texttt{4.05\small{ ±0.28}} & \texttt{.017\small{ ±.003}} & \texttt{\textbf{0.10}\small{ ±0.01}} & \texttt{1.02} \\
\bottomrule
\end{tabular}\caption{Performance comparison between the exact samplers on 4D task (strict versions).}
\label{tab:Exact4D}
\end{table}


\subsection{Comparison of Approximate Samplers}

\begin{figure}
    \includegraphics[width=1\linewidth]{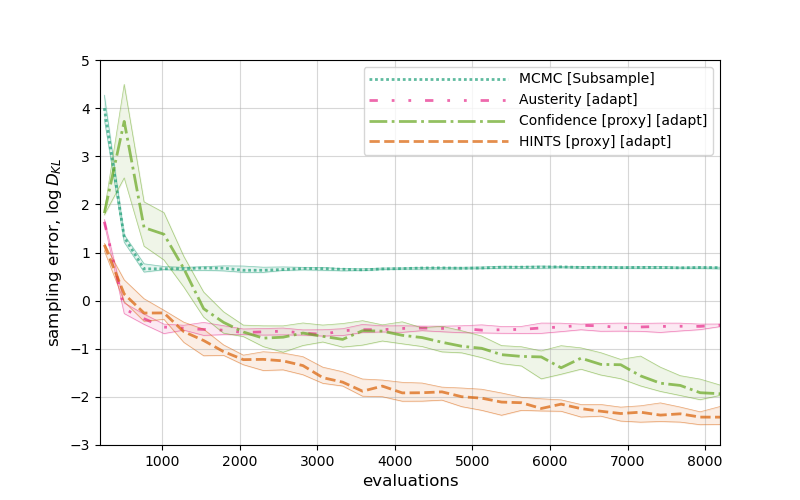}
    \caption{Comparison of sampling errors between approximate samplers on a 4D task.}
    \label{fig:Approx4DError}
\end{figure}

Both the Austerity and Confidence samplers are approximate by design, while the non-strict HINTS (in which proxy and adaptation are not frozen) might also not provide an exact sample, as discussed in \Cref{subsec:adaptive_MCMC}. In \Cref{fig:Approx4DError}, Subsample MCMC, which makes every decision on a subset of fixed size, leads to a very poor sample. The Austerity sampler, by taking larger subsets when the $t$ statistic is not significant, achieves a lower asymptotic error, but this is clearly poorer than the Confidence and HINTS samplers that continue to show improvement as more iterations are taken. The non-strict variant of HINTS achieves a slightly better final error than the exact version, suggesting that mixing performance is more important than strict reversibility on this challenging task (\Cref {tab:Approx4D}).

\begin{table}
\begin{tabular}{lrrrrr}
\toprule
\fbox{4D} & \multicolumn{1}{c}{accept/eval} & \multicolumn{1}{c}{variance/eval} & \multicolumn{1}{c}{ESS/eval} & \multicolumn{1}{c}{$D_{KL}$} & \multicolumn{1}{c}{$\hat{R}$} \\
\midrule
MCMC & \texttt{0.06\small{ ±0.01}} & \texttt{0.78\small{ ±0.07}} & \texttt{.007\small{ ±.001}} & \texttt{0.23\small{ ±0.03}} & \texttt{1.04} \\
Subsample MCMC & \texttt{2.23\small{ ±0.01}} & \texttt{8.72\small{ ±0.05}} & \texttt{.012\small{ ±.002}} & \texttt{1.97\small{ ±0.03}} & \texttt{1.01} \\
Firefly & \texttt{0.07\small{ ±0.01}} & \texttt{0.88\small{ ±0.09}} & \texttt{.008\small{ ±.002}} & \texttt{0.23\small{ ±0.03}} & \texttt{1.02} \\
HINTS & \texttt{0.05\small{ ±0.00}} & \texttt{0.71\small{ ±0.05}} & \texttt{.008\small{ ±.001}} & \texttt{0.24\small{ ±0.02}} & \texttt{1.03} \\
HINTS+proxy & \texttt{0.13\small{ ±0.01}} & \texttt{4.05\small{ ±0.33}} & \texttt{.016\small{ ±.003}} & \texttt{\textbf{0.09}\small{ ±0.02}} & \texttt{1.01} \\
Austerity & \texttt{0.19\small{ ±0.02}} & \texttt{8.92\small{ ±0.28}} & \texttt{.032\small{ ±.004}} & \texttt{0.60\small{ ±0.02}} & \texttt{1.00} \\
Confidence+proxy & \texttt{0.07\small{ ±0.01}} & \texttt{0.94\small{ ±0.08}} & \texttt{.007\small{ ±.001}} & \texttt{0.16\small{ ±0.02}} & \texttt{1.03} \\
\bottomrule
\end{tabular}
\caption{Performance comparison between the approximate samplers on 4D task.}
\label{tab:Approx4D}
\end{table}

\FloatBarrier

\subsection{High Correlation Case}\label{subsec:corr}

Here we use a modified version of the 4D test problem, with increased correlation between consecutive state dimensions, causing the total likelihood to have a narrow ridge of high likelihood. Our expectation is that the random walk proposal will have a lower acceptance rate because only proposals along the direction of the ridge will be accepted. \Cref{tab:Exact4Dcorr} confirms that the directed proposals constructed by HINTS (even without a proxy) are advantageous on this task. (HINTS is the only sampler that constructs composite proposals by combining multiple random walk proposals accepted on subsets.)

\begin{table}
\centering
\begin{tabular}{lrrrrr}
\toprule
\fbox{4D (correl.)} & \multicolumn{1}{c}{accept/eval} & \multicolumn{1}{c}{variance/eval} & \multicolumn{1}{c}{ESS/eval} & \multicolumn{1}{c}{$D_{KL}$} & \multicolumn{1}{c}{$\hat{R}$} \\
\midrule
MCMC & \texttt{0.05\small{ ±0.01}} & \texttt{3.64\small{ ±0.25}} & \texttt{.004\small{ ±.000}} & \texttt{1.43\small{ ±0.30}} & \texttt{1.21} \\
Firefly & \texttt{0.09\small{ ±0.01}} & \texttt{3.91\small{ ±0.23}} & \texttt{.004\small{ ±.000}} & \texttt{2.42\small{ ±0.41}} & \texttt{1.69} \\
HINTS & \texttt{0.06\small{ ±0.00}} & \texttt{7.96\small{ ±0.69}} & \texttt{.005\small{ ±.001}} & \texttt{0.51\small{ ±0.13}} & \texttt{1.10} \\
HINTS+proxy & \texttt{0.23\small{ ±0.00}} & \texttt{53.14\small{ ±2.17}} & \texttt{.010\small{ ±.001}} & \texttt{\textbf{0.15}\small{ ±0.03}} & \texttt{1.02} \\
\bottomrule
\end{tabular}\caption{Performance comparison for 4D task with increased correlation.}
\label{tab:Exact4Dcorr}
\end{table}

\begin{figure}
    \centering
    \subfloat[MCMC sample]{\includegraphics[width=0.5\textwidth]{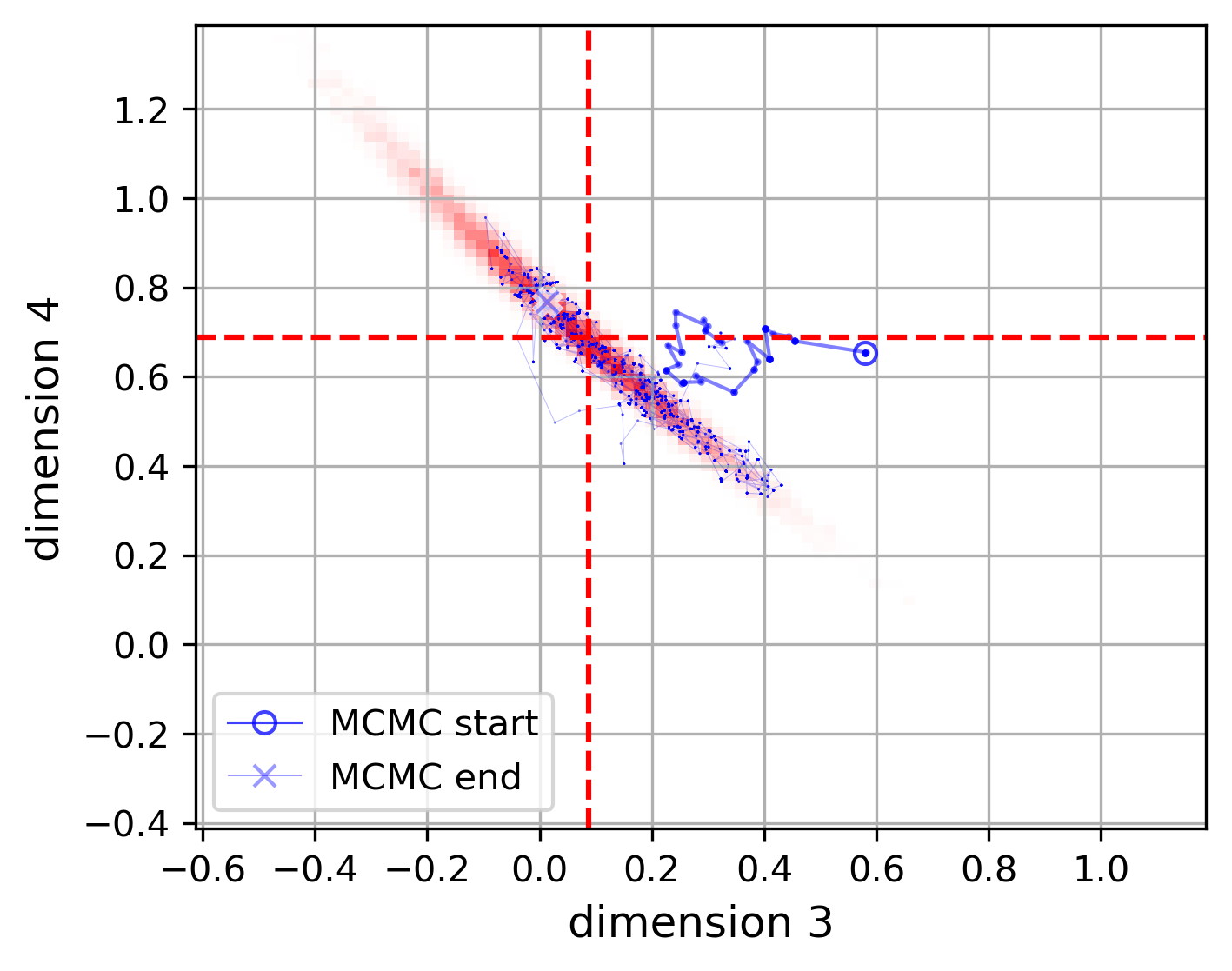}}
    \subfloat[HINTS (with proxy) sample]{\includegraphics[width=0.5\textwidth]{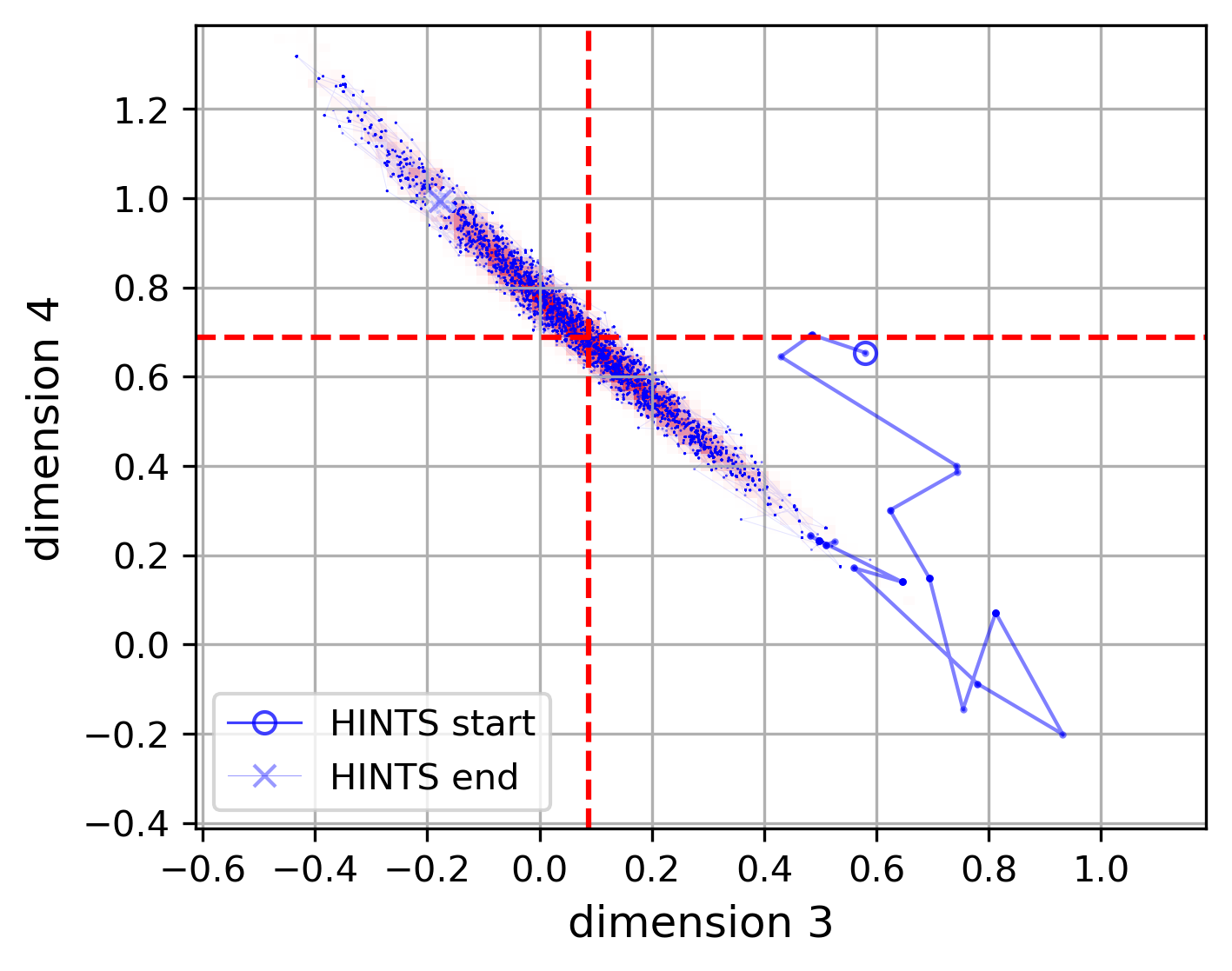}}
    \caption{2D marginal of 4D high-correlation task: the reference sample is binned in red; sample chains are shown in blue with a thicker line for first 50 iterations.}
\label{fig:chains}
\end{figure}

For a more detailed view, \Cref{fig:chains} shows the comparison between MCMC and HINTS sample chains. As this is a 4D task, a 2D projection is shown, but results were consistent for other pairs of state dimensions. MCMC is slow to find the high density ridge,  making little progress during the first 50 iterations (thicker line). It then fails to explore the whole distribution because it has to take small steps to stay inside the ridge. HINTS, given the same computation budget, explores the whole ridge with its large adapted step size; the cheap proxy evaluations filter out proposals that `fall off' the ridge.

We also expected that Firefly---and other samplers that \emph{trust} their proxy or bound---could suffer in this task because there is curvature in the ridged peak, and so the quadratic fit will be poor. However, when a quadratic proxy was used with HINTS it provided a significant improvement in all metrics, despite the curvature. In a comparison (not shown) that included approximate methods, the Austerity sampler (which does not use a proxy) achieved similar final performance to HINTS (without proxy), but HINTS was the only sampler to make effective use of a proxy. Being a delayed acceptance approach, HINTS always has the option to reject bad proxy-derived proposals using the full likelihood: we observe the curvature is also present in the generated sample.

\subsection{Quadratic Proxy Breakdown}\label{subsec:breakdown}

Despite the robustness of HINTS to some non-Gaussian structure, we expect performance of all samplers to degrade once the corresponding MVN described in \Cref{subsec:proxies} becomes a poor importance sampler for the true target. For example, a proxy that incorrectly assigns near-zero likelihood to some region of the target density can be expected to cause very slow mixing. This non-Gaussian structure can occur, particularly, when there is some ambiguity between model parameters (multiple possible explanations for the same data), a misspecified model (\textit{e.g.,} where the data is actually from a mixture) or small numbers of independent data items (\textit{e.g.,} only a few time series). All these are common in epidemiological modelling. 

\begin{figure}
    \centering
    \subfloat[Quadratic proxy]{\includegraphics[width=0.5\textwidth]{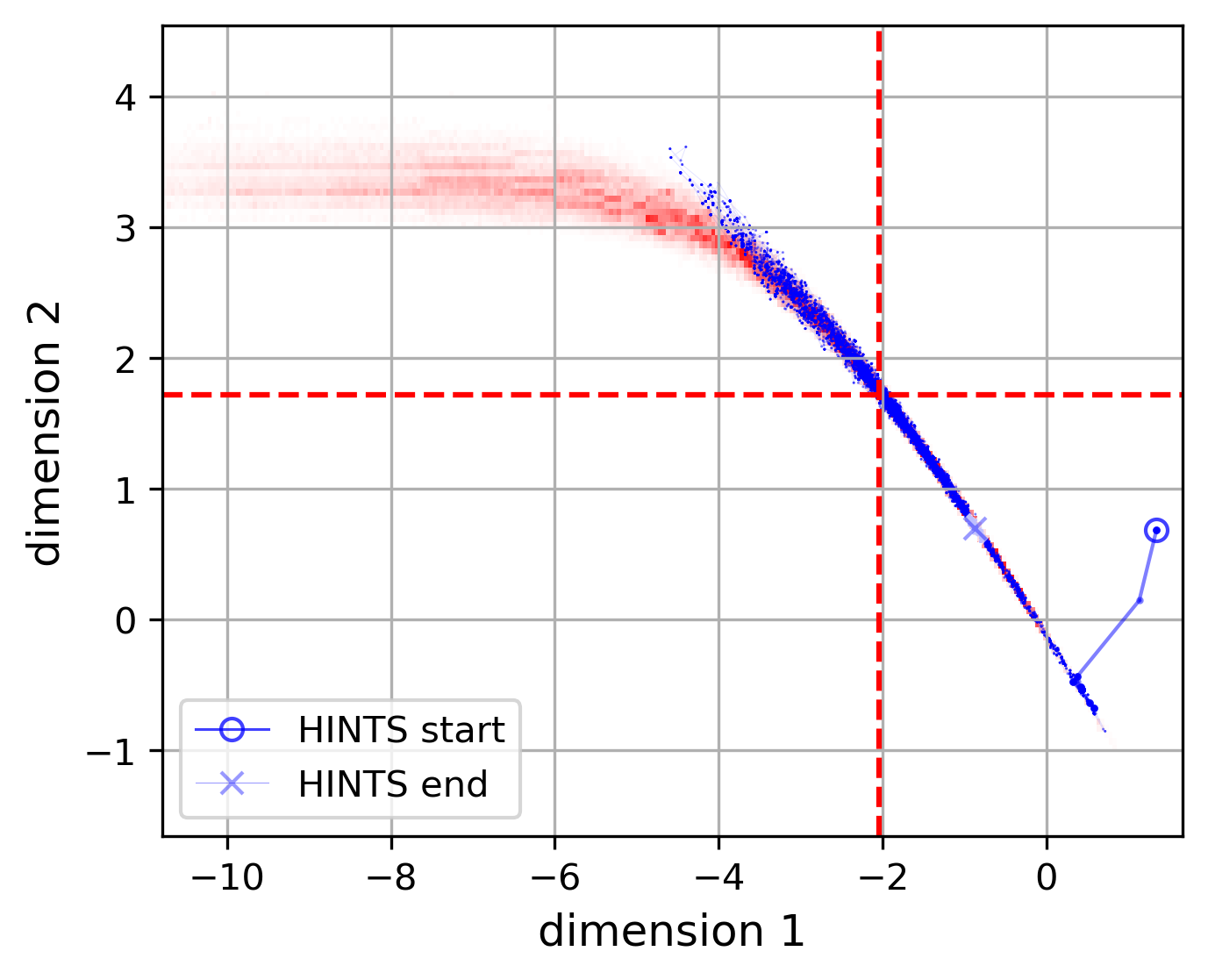}}
    \subfloat[Nearest neighbour proxy]{\includegraphics[width=0.5\textwidth]{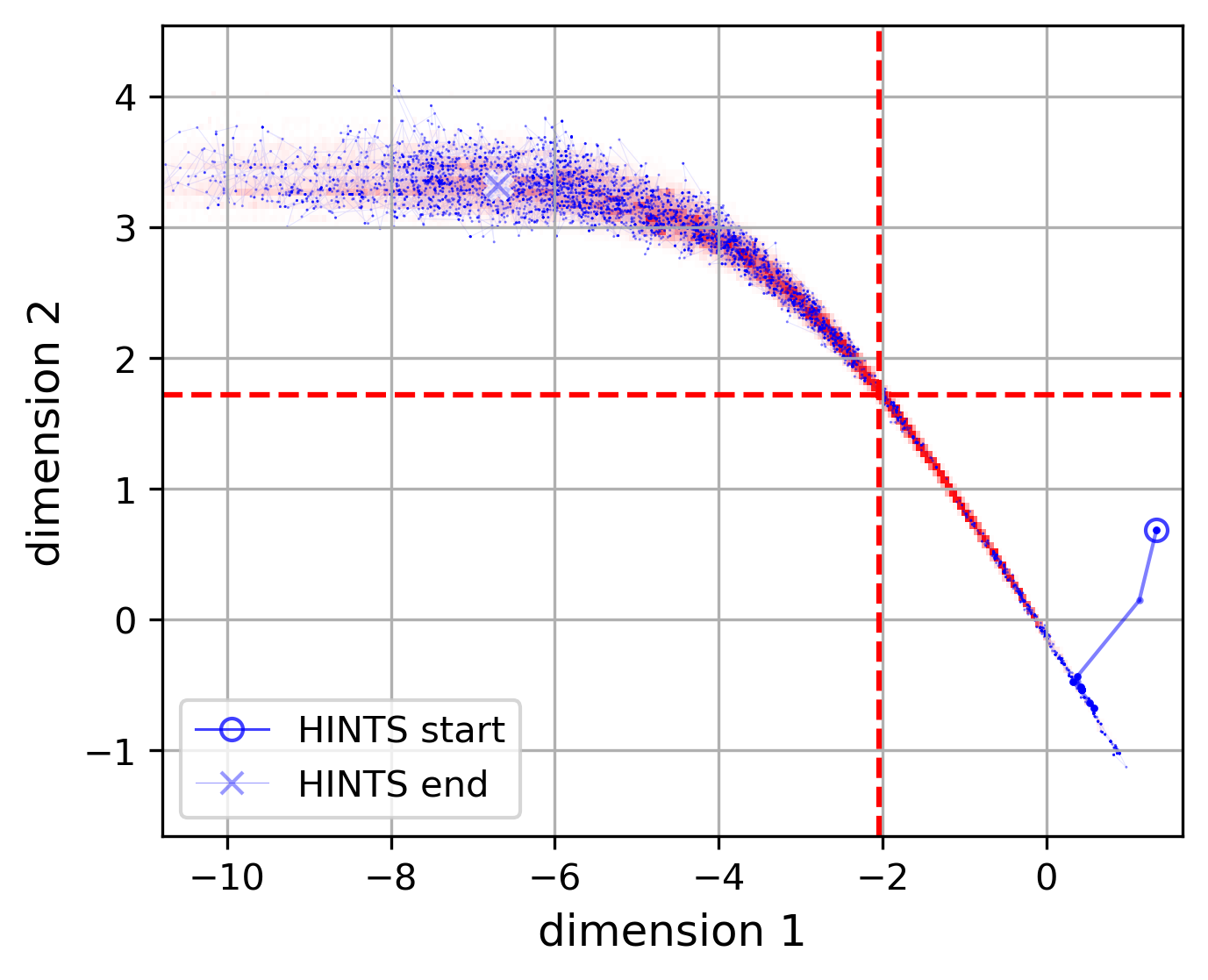}}
    \caption{In the 2D high-correlation task the target is sufficiently non-Gaussian that a better sample is obtained with a nearest neighbour proxy than with a quadratic.}
\label{fig:chains2}
\end{figure}

We now consider a 2D task with a very non-Gaussian target and compare the global quadratic proxy with the NN proxy that does not extrapolate into distant regions of state space. \Cref{fig:chains2} compares HINTS chains with each of the two proxies. It is clear that the sample obtained with the quadratic proxy is biased towards the part of the target that overlaps with an elliptical fit to the main ridge structure. (Even if the sampler is configured to provide a strict exact sample, the mixing time into the unexplored region could prevent it from doing so.) In contrast, the NN proxy mapped out the full target sufficiently quickly that the chain could provide a good sample. \Cref{tab:nearest} provides detailed metrics for multiple configurations of interest (and includes $\log D_{KL}$ to reflect the very wide range of performance). We infer:
\begin{itemize}[itemsep=0.1pt,topsep=0.1pt]
\item The quadratic proxy fails to perform well with HINTS or Confidence sampler.
\item HINTS \textit{without} a proxy is better than using a poor proxy and improves on MCMC, because it uses subset evaluations to construct directed proposals.
\item HINTS can make effective use of the NN proxy to obtain a high variance rate and low sampling error on this challenging task.
\item HINTS with a flat structure in which there are no subset evaluations performs significantly worse than our default hierarchy that has 2 levels of subsets (4 and 16 scenarios out of 64). This is important evidence that subset evaluations are useful beyond burn-in, when the proxy is not global.
\item The Confidence sampler does not make very effective use of the NN proxy. We speculate that this is because it requires proxy evaluations on all scenarios that are not in the current subset, which incur \textit{actual} likelihood evaluations for proposals beyond the explored region; \textit{i.e.,} partial proxies are not so useful in the Confidence sampler. 
\end{itemize}

\begin{table}[t!]
\centering
\begin{tabular}{lrrrrr}
\toprule
\fbox{2D (correl.)} & accept/eval & variance/eval & ESS/eval & $\log D_{KL}$ & $\hat{R}$ \\
\midrule
MCMC & \texttt{0.14\small{ ±0.01}} & \texttt{13.60\small{ ±1.70}} & \texttt{.0022\small{ ±.0002}} & \texttt{6.62\small{ ±0.47}} & \texttt{2.04} \\
HINTS (full) & \texttt{0.13\small{ ±0.01}} & \texttt{27.09\small{ ±8.46}} & \texttt{.0021\small{ ±.0002}} & \texttt{4.52\small{ ±1.90}} & \texttt{1.99} \\
HINTS (full)+Quad & \texttt{0.21\small{ ±0.01}} & \texttt{84.20\small{ ±3.78}} & \texttt{.0045\small{ ±.0012}} & \texttt{6.85\small{ ±0.18}} & \texttt{1.34} \\
HINTS (flat)+Near & \texttt{0.10\small{ ±0.01}} & \texttt{11.96\small{ ±2.68}} & \texttt{.0020\small{ ±.0002}} & \texttt{6.06\small{ ±0.76}} & \texttt{2.16} \\
HINTS (full)+Near & \texttt{0.16\small{ ±0.01}} & \texttt{43.23\small{ ±10.0}} & \texttt{.0026\small{ ±.0003}} & \texttt{\textbf{3.19}\small{ ±0.43}} & \texttt{1.96} \\
Confidence+Quad & \texttt{0.14\small{ ±0.02}} & \texttt{11.00\small{ ±1.88}} & \texttt{.0022\small{ ±.0003}} & \texttt{7.41\small{ ±0.55}} & \texttt{2.28} \\
Confidence+Near & \texttt{0.16\small{ ±0.03}} & \texttt{15.59\small{ ±4.96}} & \texttt{.0021\small{ ±.0001}} & \texttt{6.24\small{ ±0.99}} & \texttt{2.54} \\
\bottomrule
\end{tabular}
\caption{Performance comparison for differing proxies and hierarchies.}
\label{tab:nearest}
\end{table}

\subsection{Tall Data}

\begin{figure}
    \centering
    \includegraphics[width=0.8\linewidth]{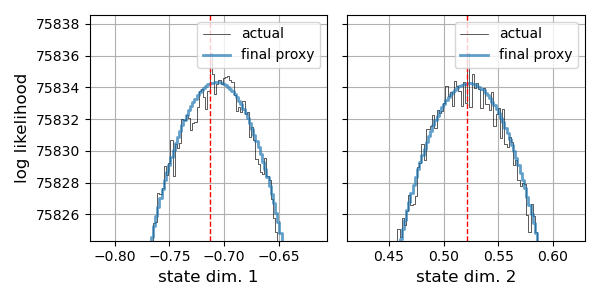}
    \caption{Proxy for 2D task with tall data: the quadratic structure is clear.}
    \label{fig:ProxyTall}
\end{figure}

In this version of the task, data set size was increased to 64 batches of 250 measurements, with each measurement having reduced information content (\Cref{app:synthetic}). With more data we expect the quadratic proxy to be a good fit (closer to the BvM bound), confirmed by \Cref{fig:ProxyTall}. Consistent with this \Cref{tab:Tall} shows that quadratic proxy-based samplers had no difficulty with this task, but Firefly was not able to establish a tight lower bound, because noise levels remain significant when summed over a large data set.

\begin{table}[h]
\centering
\begin{tabular}{lrrrrr}
\toprule
\fbox{2D (tall)} & \multicolumn{1}{c}{accept/eval} & \multicolumn{1}{c}{variance/eval} & \multicolumn{1}{c}{ESS/eval} & \multicolumn{1}{c}{$D_{KL}$} & \multicolumn{1}{c}{$\hat{R}$} \\
\midrule
MCMC & \texttt{0.29\small{ ±0.01}} & \texttt{1.54\small{ ±0.10}} & \texttt{.070\small{ ±.006}} & \texttt{0.022\small{ ±.006}} & \texttt{1.007} \\
Subsample MCMC & \texttt{2.44\small{ ±0.01}} & \texttt{4.48\small{ ±0.05}} & \texttt{.062\small{ ±.004}} & \texttt{0.498\small{ ±.019}} & \texttt{1.011} \\
Firefly & \texttt{0.33\small{ ±0.04}} & \texttt{1.79\small{ ±0.17}} & \texttt{.090\small{ ±.014}} & \texttt{0.025\small{ ±.007}} & \texttt{1.061} \\
HINTS & \texttt{0.18\small{ ±0.00}} & \texttt{1.27\small{ ±0.09}} & \texttt{.072\small{ ±.009}} & \texttt{0.034\small{ ±.010}} & \texttt{1.014} \\
HINTS+proxy & \texttt{0.65\small{ ±0.00}} & \texttt{7.52\small{ ±0.37}} & \texttt{.286\small{ ±.026}} & \texttt{\textbf{0.007}\small{ ±.002}} & \texttt{1.002} \\
Austerity & \texttt{0.30\small{ ±0.02}} & \texttt{4.97\small{ ±0.34}} & \texttt{.112\small{ ±.014}} & \texttt{0.212\small{ ±.018}} & \texttt{1.005} \\
Confidence+proxy & \texttt{0.27\small{ ±0.03}} & \texttt{1.66\small{ ±0.09}} & \texttt{.076\small{ ±.007}} & \texttt{0.024\small{ ±.005}} & \texttt{1.008} \\
\bottomrule
\end{tabular}
\caption{Performance comparison for tall data.}
\label{tab:Tall}
\end{table}

When we extended the high-correlation task of \Cref{subsec:breakdown} to tall data, similar results were obtained: the NN proxy was again required in place of the quadratic, subset evaluations were valuable, and only HINTS could provide an effective sample.

\subsection{Higher Dimensions}
\begin{figure}
    \centering
    \includegraphics[width=1\linewidth]{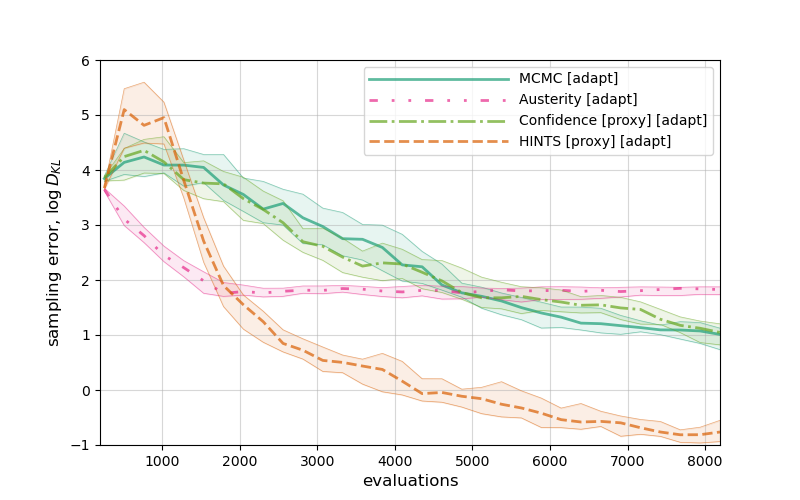}
    \caption{Comparison of sampling errors for 16D task; see table for other samplers.}
    \label{fig:Comparison16D}
\end{figure}

Using MCMC methods in high-dimensional state spaces is challenging especially where gradient information cannot be used: it is difficult for a sampling chain to traverse the high-dimensional space using random walk proposals. We found that most methods failed when dimensionality of our test problem was increased to 16 (\Cref{tab:16D}), but HINTS with a proxy was able to achieve a good level of accuracy ($D_{KL} = 0.48$) in the 8192 evaluations budget (\Cref{fig:Comparison16D}). Full adaptive MCMC was still making progress at this stage, while HINTS without proxy and the Confidence sampler with a proxy showed no advantage over MCMC. Meanwhile, the Austerity sampler and Firefly still continued to exhibit very large errors. Notably, the Austerity sampler has fast burn-in performance and could be useful for finding a (unimodal) high-density region before switching to a more accurate sampler. All methods failed when extra correlation between state dimensions was added. Therefore, it is very important to remove any known redundancy prior to sampling, and it may be necessary to factor the state space to make progress in such cases.

\begin{table}[h]
\centering
\begin{tabular}{lrrrrr}
\toprule
\fbox{16D} & \multicolumn{1}{c}{accept/eval} & \multicolumn{1}{c}{variance/eval} & \multicolumn{1}{c}{ESS/eval} & \multicolumn{1}{c}{$D_{KL}$} & \multicolumn{1}{c}{$\hat{R}$} \\
\midrule
MCMC & \texttt{0.06\small{ ±0.01}} & \texttt{1.04\small{ ±0.11}} & \texttt{.004\small{ ±.000}} & \texttt{2.64\small{ ±0.44}} & \texttt{1.09} \\
Subsample MCMC & \texttt{2.06\small{ ±0.01}} & \texttt{32.76\small{ ±0.12}} & \texttt{.004\small{ ±.000}} & \texttt{14.95\small{ ±0.30}} & \texttt{1.05} \\
Firefly & \texttt{0.05\small{ ±0.02}} & \texttt{4.24\small{ ±3.26}} & \texttt{.004\small{ ±.000}} & \texttt{32.72\small{ ±4.64}} & \texttt{2.02} \\
HINTS & \texttt{0.04\small{ ±0.01}} & \texttt{0.83\small{ ±0.10}} & \texttt{.004\small{ ±.000}} & \texttt{3.45\small{ ±0.67}} & \texttt{1.63} \\
HINTS+proxy & \texttt{0.14\small{ ±0.01}} & \texttt{6.88\small{ ±1.35}} & \texttt{.006\small{ ±.001}} & \texttt{\textbf{0.48}\small{ ±0.09}} & \texttt{1.09} \\
Austerity & \texttt{0.14\small{ ±0.01}} & \texttt{16.73\small{ ±1.38}} & \texttt{.005\small{ ±.001}} & \texttt{6.20\small{ ±0.45}} & \texttt{1.02} \\
Confidence+proxy & \texttt{0.05\small{ ±0.01}} & \texttt{0.96\small{ ±0.10}} & \texttt{.004\small{ ±.000}} & \texttt{2.80\small{ ±0.53}} & \texttt{1.10} \\
\bottomrule
\end{tabular}
\caption{Performance comparison for 16D task.}
\label{tab:16D}
\end{table}

\subsection{Disease Task Results}\label{subsec:disease_results}

This is the most representative task for realistic PMCMC noise characteristics because each likelihood evaluation is from a full PF run. \Cref{tab:Disease} shows results that are consistent with the synthetic examples. Again, Subsample MCMC and Austerity samplers show large errors. Confidence and Firefly show no benefit over full MCMC. HINTS with a quadratic proxy achieves better mixing and lower error than the other samplers. (The NN proxy was not necessary.) This consistency provides supporting evidence that results earlier in this section would be representative in full particle MCMC.

\begin{table}[h]
\centering
\begin{tabular}{lrrrrr}
\toprule
\fbox{Disease} & \multicolumn{1}{c}{accept/eval} & \multicolumn{1}{c}{variance/eval} & \multicolumn{1}{c}{ESS/eval} & \multicolumn{1}{c}{$D_{KL}$} & \multicolumn{1}{c}{$\hat{R}$} \\
\midrule
MCMC & \texttt{0.20\small{ ±0.01}} & \texttt{0.48\small{ ±0.01}} & \texttt{.047\small{ ±.005}} & \texttt{0.010\small{ ±.002}} & \texttt{1.002} \\
Subsample MCMC & \texttt{2.20\small{ ±0.00}} & \texttt{3.93\small{ ±0.01}} & \texttt{.035\small{ ±.003}} & \texttt{1.081\small{ ±.024}} & \texttt{1.004} \\
Firefly & \texttt{0.24\small{ ±0.01}} & \texttt{0.53\small{ ±0.01}} & \texttt{.047\small{ ±.006}} & \texttt{0.009\small{ ±.002}} & \texttt{1.003} \\
HINTS & \texttt{0.13\small{ ±0.01}} & \texttt{0.41\small{ ±0.02}} & \texttt{.043\small{ ±.004}} & \texttt{0.010\small{ ±.002}} & \texttt{1.003} \\
HINTS+proxy & \texttt{0.47\small{ ±0.00}} & \texttt{2.32\small{ ±0.05}} & \texttt{.176\small{ ±.009}} & \texttt{\textbf{0.004}\small{ ±.001}} & \texttt{1.001} \\
Austerity & \texttt{0.22\small{ ±0.01}} & \texttt{2.56\small{ ±0.13}} & \texttt{.078\small{ ±.004}} & \texttt{0.411\small{ ±.025}} & \texttt{1.001} \\
Confidence+proxy & \texttt{0.21\small{ ±0.00}} & \texttt{0.49\small{ ±0.02}} & \texttt{.043\small{ ±.006}} & \texttt{0.009\small{ ±.001}} & \texttt{1.003} \\
\bottomrule
\end{tabular}
\caption{Performance comparison for 2D disease task.}
\label{tab:Disease}
\end{table}

%
%
%
%
\FloatBarrier

\section{Discussion and Related Work}
\label{sec:discussion}

Here, we discuss the research progress we have made in a broader context, including providing some more insight as to why some sampling approaches are successful and others fail on challenging tasks. 

\subsection{Computation-Aware Adaptive Control}

Our approach to adaptation within MCMC is the first we are aware of that takes into account the computational cost of different decisions. We have seen that this avoids a preliminary search over proposal parameters and opens up new regimes that achieve faster mixing than acceptance rate targets that are not universal. \citet{everitt2008montecarlo} explored using the product of autocorrelation time and computation per step as an optimisation objective for selecting between alternative sampling hierarchies in HINTS. \citet{sherlock2015efficiency} and \citet{doucet2015efficient} explored the trade-off between the number of unbiased estimates (\textit{e.g.,} particles) computed per likelihood evaluation and the progress of a pseudo-marginal MCMC chain, although not adaptively. 

The scalar parameter controlled by our adaptive algorithm is not limited to simple Gaussian proposals but can be used with other densities or as a parameter for specialised proposals, like discrete kernels. Furthermore, the action set can be extended to select between different proposal types as well as different scale factors.

While computation-aware control is particularly relevant for subset-based samplers, it is more broadly applicable: whenever there is variation in computational cost between the proposal decision options, we expect this approach to outperform existing methods. In delayed acceptance methods, a proposal that is rejected using a cheap proxy consumes less computational resource. In samplers that mix different proposal types, some may not require as much computation; \textit{e.g.,} for a mixture model, proposals that only affect the parameters of one component will be evaluated only on the relevant data. In decision tree sampling, the computational cost is dependent on both the proposal type and the location in the tree. 

For inexact methods (Austerity and Confidence samplers), there could be a dependency between the proposal scale and the asymptotic accuracy of the sampler. Our adaptation mechanism simply tries to maximise the variance per unit of computation for the chain's progress; there is no guarantee that the impact of the occasional incorrect decisions made by these samplers will not be worse at the optimised proposal scale. Other adaptive MCMC approaches have the same problem, and this provides further motivation to prefer exact samplers. 

\subsection{Proxies}

We adapted existing samplers (Firefly, Confidence, HINTS) to use data-driven proxies that avoid  analytical derivatives. The quadratic form provided a good interpolation function to the overall target density when working with many scenarios, tall data, or noise-free test problems. The quadratic proxy has $O(D^2)$ degrees of freedom in the dimension $D$ of the state space, and we were able to use this successfully in up to 16 dimensions. In higher dimensions we expect that some regularisation would be required; for example through restricting to a diagonal covariance structure. Ongoing sampling after establishing the quadratic proxy in any of the samplers could probably be simplified by switching to a single-level delayed-acceptance structure, with cheap composite proposals constructed from a chain of full proxy evaluations (a special case of HINTS).

At the subset level (\textit{e.g.,} a few scenarios) the quadratic fits for the irregular log likelihoods are misleading, and this is a likely cause of the failure (often) of the Confidence sampler to outperform full MCMC. The irregular likelihoods were even more problematic for obtaining a tight quadratic \emph{lower bound} in Firefly. HINTS was more robust to irregular and non-Normal likelihoods, but very high correlation in state space caused it to fail to benefit from the quadratic proxy.

When the likelihood is non-Normal, aggressively dropping proxies so as to obtain \emph{local} quadratic fits is feasible and was proposed for the Confidence sampler when individual Taylor expansions were found to offer poor global approximations \citep{Bardenet2017}. However, for strict exact sampling, the proxy must be frozen for the sampling interval (to achieve reversibility), and so a global proxy that works well is much more desirable. For a data-driven proxy this could be achieved by fitting a mixture of Gaussians in place of the single MVN.

When the global quadratic proxy was not appropriate, we found that a NN proxy could be effective. We enforced a distance constraint so that the proxy would not extrapolate to distant states. This allows the chain to achieve a high variance rate (per actual evaluation) in explored regions while still making use of information from actual likelihood subset evaluations for distant proposals. We envisage that the NN proxy would be less useful in higher dimensions and that proxies may therefore have to be designed on a task-specific basis.

\subsection{Samplers}

We chose to include Firefly \citep{maclaurin2014} as representative of a class of samplers that exploit known bounds on likelihood. \citet{pmlr-v97-cornish19a} proposed Scalable Metropolis-Hastings (SMH), using a \emph{local} constraint on the change in the target between two states in a Factorised Metropolis-Hastings (FMH) formulation. \cite{zhang2020asymptotically} proposed TunaMH, which also depends on strong local bounds on the target to identify a batch size that yields exact samples, also without the need for auxiliary variables. \cite{Prado2024} introduced Metropolis–Hastings with Scalable Subsampling (MH-SS), which makes use of model-specific likelihood bounds (\textit{e.g.,} for Generalised Linear Models) to obtain an exact sampler with embedded subset evaluations. 
Subset evaluations cannot normally be used directly in a pseudo-marginal sampling chain because they are unbiased in $\log$ likelihood but biased in likelihood. However, \citet{quiroz2018speeding} was able to bound approximation error when good control variates were available, yielding very competitive results for \textit{regular} likelihoods.

In all these approaches, to obtain a speed-up compared with full MCMC requires the global or local bounds to be tight. For our irregular likelihoods that contain fat-tailed noise (\textit{e.g.,} localised spikes) we have only a data-driven bound, and we have seen this is not tight, causing this type of approach to degrade, especially as dimensionality grows. In contrast, HINTS can exploit an \emph{interpolating proxy} that is robust to localised noise spikes. 

In some tall data sampling tasks a subset of the data called a `coreset' can be inferred that approximates the full data sufficiently well, so that the scaled coreset log likelihood can be used in place of full evaluations by an approximate sampling algorithm \citep{pmlr-v238-chen24f}. This can be useful if there is an imbalance in the information content between the likelihood functions for different data points (or scenarios), or if there is a requirement to generate samples with a reduced memory footprint. In the absence of these characteristics, the inclusion of a preliminary (or parallel) coreset inference task is not a natural approach compared with revisiting all the data periodically. We also note that the most advanced coreset methods make use of gradient information that is assumed unavailable here. However, for tasks where there is known to be an information imbalance between scenarios, the same criteria used in coreset methods might be useful to weight the data for non-uniform subset sampling.

HINTS has been largely overlooked in the sampling literature, perhaps because it was introduced as a reinforcement learning algorithm, and without proof that it is an exact sampler. Essentially, HINTS builds directed proposals from cheap subset evaluations that allow the (expensive) top-level sampling chain to take larger moves \textit{and} achieve high acceptance rates. We have presented a proof that HINTS is an exact sampler for the target density, at the root-node output. This proof is consistent with HINTS viewed as a hierarchical generalisation of \emph{delayed acceptance}.

Subsequently, Multi-Level MCMC \citep{Dodwell2015} and particularly `Multi Level Delayed Acceptance' (MLDA) \citep{Lykkegaard_2023} also use delayed acceptance in a multi-level hierarchy with sub-chains. The difference is that HINTS uses subset evaluations to obtain the `coarse' (cheap to compute) evaluations, while MLDA uses a factorisation of the state space. Specifically, MLDA makes separate proposals at coarse and fine scales, while HINTS constructs its top-level proposals from combinations of moves made using subset evaluations. \cite{Everitt02012021} explored the use of delayed acceptance MCMC in the context of Sequential Monte Carlo (SMC) for Approximate Bayesian Computation: a `cheap simulator' (for the likelihood evaluation) was used to find potentially good proposals for a full simulation run.

Unlike the HINTS formulation in \citet{Strens2004}, we do not require that the subset log-likelihoods combine additively as we move up the tree. All that is required is that the user provides a function which extracts useful information from the subset, to inform a directed proposal for the next level up. Consider, for example, the important epidemiological disease-modelling case where we wish to run PMCMC on \emph{a single very long time series}. Despite there being only one data item, HINTS can operate with scenario evaluations that are PF runs on segments of the time series. The sum of PF log likelihoods for a string of sub-sequences provides the root-node (that does run the PF on the whole sequence) with highly informative composite proposals.  Another alternative decomposition of PMCMC processing is for each scenario to represent a cheap run with a small number of particles in the filter, providing highly directed proposals for a larger (more accurate) evaluation at the HINTS root-node. 

For all samplers, we used the same \textit{deterministic} target density built by using fixed random seeds for the stochastic likelihood evaluations (\textit{e.g.,} for the PF runs). This helped us distinguish between exact and approximate samplers, and allowed us to precompute some of the models to support larger research comparisons. However, in practical application the random seeds could be changed from time to time to operate the top-level chain in the pseudo-marginal framework, eliminating the residual biases associated with fixed seeds. Furthermore we used 50 independent runs to gather statistics for statistical comparison of samplers, whereas a better final sample would be obtained by combining samples across runs.

\subsection{Subset Evaluations and Tempering}
\label{subsec:tempering}

For awkward target densities (noisy, multi-modal or high-dimensional), running interacting chains at different temperatures has been shown to improve exploration and mixing, because the higher-temperature chains provide mobility across low-probability regions \citep{earl2005parallel}. In HINTS, additional tempering can be introduced by dividing the log likelihood at level $h$ by a temperature $T_h$: higher temperatures yield `lower-contrast' versions of the likelihood. Unbiased samples will always be achieved at the root ($h = H$) if the temperature at that level is 1. \citet{Strens2004} obtained optimisation instead of sampling by tempering the HINTS sampling with a root-node temperature close to zero.

However, in Bayesian inference with an additive log likelihood decomposition, there is a `natural' tempering effect: the log likelihood sums for subsets have smaller variation than the full sum, and we have not found it necessary to move away from this. Nevertheless, the natural tempering could explain why HINTS is successful on challenging targets: proposals accepted with subset evaluations (using actual likelihoods, or proxies) provide mobility for exploring the full target that has larger likelihood variation.

\section{Conclusions \& Future Work}
\label{sec:Conclusions}

This work has explored subset samplers for Bayesian inference with irregular and computationally expensive likelihoods.  \Cref{tab:improved} highlights the modifications we have made and evaluated to extend existing subset methods (\Cref{tab:original}) for this setting.

We needed methods that are accelerated by the use of a proxy (or known bounds) but do not require analytical derivatives, because these are not available for irregular likelihoods. Therefore, we integrated data-driven proxies into several samplers. A global per-scenario quadratic proxy has the correct functional form in the BvM limit, while a NN proxy, with partial coverage of the state space in a trusted region, was proposed for non-Normal targets.

We developed a computation-aware adaptive algorithm in order to extract maximum performance from MCMC samplers where each proposal has a different expected computational cost. This has helped us operate each sampler more effectively on tasks where the likelihood evaluations dominated the computational cost. 

\begin{table}
\centering
\begin{tabular}{lccc}
\toprule
\textbf{Sampler}& \textbf{Exact/Approximate} & \textbf{Proxy or bound} & \textbf{Adaptive} \\
\midrule 
Full MCMC & Exact & No & Yes \\
Subsample MCMC & Approximate & No & \multirow{4}{*}{\rotatebox[]{90}{\fbox{\textbf{Cost-aware}}}} \\
Austerity   & Approximate & No & \\
Confidence & Approximate & \textbf{Data driven} & \\
Firefly & Exact & \textbf{Data driven} bound  & \\
HINTS& Exact \textbf{(proof)}  or approx. & \textbf{Data driven} & \\     
\bottomrule 
\end{tabular}
\caption{List of subset samplers with modifications and improvements in \textbf{bold}.}
\label{tab:improved}
\end{table}

The HINTS framework leaves the user with choices about the hierarchical decomposition of the likelihood. We provided a universal configuration (3 levels with a branch factor of 4) that was successful for all tasks. We introduced downsampling, when not using a global proxy, to avoid evaluating every scenario on each cycle. Our proof that HINTS is an exact sampler is valid not just for additive likelihoods but for the general case in which subset evaluations are user-specified functions that extract useful information at reduced computational cost. \citet{Lykkegaard_2023} also proposed that lower-level chains in the sampling hierarchy for MLDA could use a randomised number of steps (Randomised-Length-Subchain Surrogate Transition algorithm). This may also be useful in HINTS to provide a wider variety of proposals to the top-level chain, particularly if the proxy is inaccurate in parts of the state space.

A large experimental evaluation of multiple sampler configurations was undertaken. This included a disease model with full PF likelihood evaluations and a flexible, fast-to-evaluate synthetic model with similar noise characteristics. We found the Subsample MCMC and Austerity \emph{approximate samplers} yielded inaccurate samples. The Confidence sampler (with proxy) was more accurate but often degraded to the performance of full MCMC on challenging tasks with irregular likelihoods.

The \textit{original} HINTS algorithm outperformed full MCMC only when there was substantial correlation between state dimensions. But when a proxy was integrated, HINTS showed much faster mixing than other samplers with the disease model and the extended tasks. Freezing both the proxy and the adaptive proposals for the final interval gives an exact HINTS sampler without a significant impact on performance.

We established that subset evaluations are valuable in the context of irregular likelihoods:
\begin{itemize}
    \item When operating without a proxy, HINTS (exact) and Austerity (inexact) samplers both make effective use of subset evaluations.
    \item With the quadratic proxy, HINTS (exact) and Confidence (inexact) samplers are both able to achieve better performance than full MCMC on tasks where the target is near-Normal. HINTS was more robust to deviation from Normal. In this case, once a cheap and accurate proxy is established, we verified it is possible to use full-proxy sub-chains, without any subset evaluations.
    \item For highly non-Normal tasks we used a NN proxy that only provides cheap evaluations in the explored region of the target. This means subset evaluations using the actual likelihood take place for proposals into unexplored regions of state space. With this partial proxy, we found that HINTS made good use of the subset evaluations to outperform full MCMC, while the Confidence sampler gained little benefit.
\end{itemize}

Effective use of proxies is essential as tasks become more complex. The reason for HINTS outperforming other subset approaches appears to be the different way in which it combines actual evaluations with proxy evaluations: in HINTS the proxy or `cheap' surrogate is never completely trusted because it is only included in the proposal mechanism for delayed acceptance. In contrast, extending a subset evaluation with proxy estimates for the other scenarios (with Austerity and Confidence samplers) means that any inaccurate scenario proxy evaluation can damage the quality of acceptance decisions. HINTS also benefits from the natural tempering effect of subset evaluations discussed in \Cref{subsec:tempering}.

We were able to use parallelism for the innermost computations (\textit{e.g.,} PF runs) and also across the multiple runs at the outermost level, necessary for the statistical comparisons. In practical application, samples from the  parallel runs would be merged to improve accuracy. An alternative to MCMC, which can also be parallelised, is Sequential Monte Carlo (SMC) sampling \citep{del2006sequential}. Including HINTS inside an SMC sampler is an active area of research that could provide both a speed-up and better samples on large inference tasks. Furthermore, considering a time series problem in this framework would result in an alternative to SMC$^2$ \citep{chopin2013smc2, rosato2023mathcal}.

While we have focused on random walk proposals due to the lack of gradient information in our target models, HINTS allows gradient-based proposals when they are available. Therefore Langevin dynamics or HMC can be used in the subset chains. This could significantly improve sampling efficiency for regular likelihoods while still benefiting from the hierarchical decomposition. HINTS composite proposals constructed using gradient information could be integrated with SMC in the structure proposed by \citet{devlin2024no}. Investigating how the HINTS adaptive control mechanism interacts with gradient-based proposals is another exciting avenue for future research.

A natural extension of this work is applying these techniques to calibrating agent-based models (ABMs), which require multiple simulations to estimate a pseudo-likelihood by comparing simulation outputs to observed time series. A relevant example is Covasim, an ABM used to model the spread of COVID-19 \citep{kerr2021covasim}. These models are computationally expensive because they involve stochastic simulations across many agents and require repeated evaluations for parameter inference. Adapting HINTS to this setting could enable more efficient inference by incorporating hierarchical likelihood evaluations and data-driven proxies to reduce the number of full simulations needed for accurate calibration. A relevant example can be found in \citep{panovska2023machine} where they use a machine learning step to pre-screen parameter sets that are predicted to perform poorly when calibrating Covasim with ABC. This pre-screening acts in a similar way to a proxy in MCMC, allowing the model to avoid running expensive simulations on low-quality candidates.

Another promising direction is the application of these methods to phylogenetic analysis, where Bayesian inference is widely used to reconstruct evolutionary histories from genetic data \citep{drummond2007beast, nascimento2017biologist}. The likelihood functions in phylogenetics are often highly irregular and expensive to evaluate, particularly for large trees with complex substitution models. The new methods we have introduced could help reduce the computational burden associated with sampling large phylogenies, potentially enabling more scalable inference for epidemiological and evolutionary studies.


\section*{Acknowledgments and Disclosure of Funding}
The authors would like to thank Dr Richard Everitt of Warwick University for his valuable feedback. For the purpose of Open Access, the authors have applied a CC BY public copyright license to any Author Accepted Manuscript version arising from this submission. Conor Rosato was funded by the Wellcome CAMO-Net UK grant: 226691/Z/22/Z; Harvinder Lehal was funded by EPSRC Centre for Doctoral Training in Distributed Algorithms and the U.K. Government through a Research Studentship under Grant EP/S023445/1 and Simon Maskell was funded by EPSRC through the Big Hypotheses under Grant EP/R018537/1 and Dstl in collaboration with the Royal Academy of Engineering. The authors thank Dstl, UK MOD and the Royal Academy of Engineering for supporting this work. The views and conclusions contained in this paper are of the authors and should not be interpreted as representing the official policies, either expressed or implied, of the UK MOD or the UK Government.


\appendix 

\section{Disease Model}

\label{app:time_series_PF_example}

We consider the stochastic Susceptible, Exposed, Infected, Recovered, Susceptible (SEIRS) disease model, which is a variant of the widely known SIR model \citep{kermack1927contribution}. 

At $t = 0$, the total population, $N_{pop}$, are divided into four unobservable compartments. The Infected compartment starts with a random number of individuals: $\mathbf{I}_0 \sim \textrm{Uniform}(N_{pop}/20,\ldots,N_{pop}/10)$. The Exposed compartment $\mathbf{E}_0$ (if being used) is initialised in the same way. The remainder of the population is placed in the susceptible compartment: $\mathbf{S}_0 = N_{pop} - \mathbf{I}_0 - \mathbf{E}_0$ and the Recovered compartment (if in use) is initially empty: $\mathbf{R}_0 = 0$.

In order to `mix' the process before generating the exemplar time series we use as data, a random warm-up period of $t_0 \sim \textrm{Uniform}(\{0,\ldots,9\})$ steps is also specified, after which a time series of length 20 will be retained.

At each time step $t > 0$, a Binomially distributed number of individuals leave each compartment. The binomial parameters are given by:
\begin{align*}
p(\mathbf{S}\rightarrow \mathbf{E})&=1-e^{-\beta \frac{\mathbf{I}_{t-1}}{N_{pop}}},\\
p(\mathbf{E}\rightarrow \mathbf{I})&=1-e^{-\delta},\\
p(\mathbf{I}\rightarrow \mathbf{R})&=1-e^{-\gamma},\\
p(\mathbf{R}\rightarrow \mathbf{S})&=1-e^{-\epsilon},
\end{align*}
where $p(\mathbf{S}\rightarrow \mathbf{E})$, $p(\mathbf{E}\rightarrow \mathbf{I})$, $p(\mathbf{I}\rightarrow \mathbf{R})$ and $p(\mathbf{R}\rightarrow \mathbf{S})$ denote the probability of each individual leaving the susceptible, exposed, infected and recovered compartments, respectively. These probabilities are governed by rate parameters $\theta=\textrm{encode}(\beta, \delta, \gamma, \epsilon)$. The $\textrm{encode()}$ function is the element-wise inverse sigmoid and serves to map each of these positive transition rates onto $[-\inf, \inf]$ (which is simpler for random walk samplers).

Note that this is not a standard hidden Markov model because the transition probability $p(\mathbf{S}\rightarrow \mathbf{E})$ is dynamic, depending on the number already infected, $\mathbf{I}_{t-1}$. The corresponding binomial distributions are
\begin{align*}
n(\mathbf{S}\rightarrow \mathbf{E})&\sim \text{Binomial}(\mathbf{S}_{t-1}, p(\mathbf{S}\rightarrow \mathbf{E})),\\
n(\mathbf{E}\rightarrow \mathbf{I})&\sim \text{Binomial}(\mathbf{E}_{t-1}, p(\mathbf{E}\rightarrow \mathbf{I})),\\
n(\mathbf{I}\rightarrow \mathbf{R})&\sim \text{Binomial}(\mathbf{I}_{t-1}, p(\mathbf{I}\rightarrow \mathbf{R})),\\
n(\mathbf{R}\rightarrow \mathbf{S})&\sim \text{Binomial}(\mathbf{R}_{t-1}, p(\mathbf{R}\rightarrow \mathbf{S})),
\end{align*}
where $n(\mathbf{S}\rightarrow \mathbf{E})$, $n(\mathbf{E}\rightarrow \mathbf{I})$, $n(\mathbf{I}\rightarrow \mathbf{R})$ and $n(\mathbf{R}\rightarrow \mathbf{S})$ denote the total number of individuals leaving each of the four compartments.

The complete discrete, stochastic SEIRS model is therefore presented as
\begin{align*}
\mathbf{S}_{t}&=\mathbf{S}_{t-1}-n(\mathbf{S}\rightarrow \mathbf{E})+n(\mathbf{R}\rightarrow \mathbf{S}),\\
\mathbf{E}_{t}&=\mathbf{E}_{t-1}+n(\mathbf{S}\rightarrow \mathbf{E})-n(\mathbf{E}\rightarrow \mathbf{I}),\\
\mathbf{I}_{t}&=\mathbf{I}_{t-1}+ n(\mathbf{E}\rightarrow \mathbf{I})-n(\mathbf{I}\rightarrow \mathbf{R}),\\
\mathbf{R}_{t}&=\mathbf{R}_{t-1}+n(\mathbf{I}\rightarrow \mathbf{R})-n(\mathbf{R}\rightarrow \mathbf{S}).
\end{align*}

The current number of infected individuals in the population, $\mathbf{I}_t$, is observed through a noisy process representing (for example) random testing of individuals:
\begin{align}
\mathbf{y}_{t} \sim \mathrm{Poisson}\left(\mathbf{y}_{t} ;\mathbf{I}_{t}\right).
\end{align} 

Each time series is obtained by capturing these observations for 20 steps ($t_0$ to $t_0+19$) which we renumber as $y_{1:20}$ for use in the PF. 

We have presented the full 4D model here but will run experiments with a reduced 2-compartment model for computational expediency, then use a faster alternative task for sampler comparisons with higher-dimensional state spaces.

\section{Particle Filter}
\label{app:particle_filter}

Assume each item in the data set is a time series of the form $\mathbf{y}_{1:t}$. A single scenario $i$ could contain a single time series or a batch of multiple time series.  While the PF is often used to infer the latent state of a dynamic system (given a sequence of observations), here it serves to evaluate the likelihood of observing a particular $\mathbf{y}_{1:t}$ for a dynamic system model with a particular choice of parameters, $\theta$. A single run processes one time series for one choice of $\theta$, so many thousands of runs may be required when it is embedded in a sampler for $\theta$. Each of these runs involves propagating thousands of particles through multiple timesteps.

Conventionally in time series filtering, $\mathbf{x}_{1:t}$ (rather than $z$) is used for latent states and we will retain this common notation in this section only. In the context of the disease model: latent states are the unknown number of individuals in each compartment, observations are the number of individuals testing positive and model parameters $\theta$ are the transmission rates between compartments.

The PF works by propagating a collection of hypotheses (particles) for the latent state through each time step in turn. The stochastic transitions of the model are simulated for each particle, and associated weights are updated according to the likelihoods of individual observations. Resampling ensures the particle distribution does not degenerate. The PF is initialised as follows:
\begin{itemize}[noitemsep,topsep=0pt]
\item Sample $N_x$ particles, $\{\textbf{x}_0^j\}_{j=1}^{N_x}$ from a specified prior $\textbf{x}_0 \sim p(\textbf{x}_0 \mid \theta)$. $N_x = 4000$ was used in our experiments. 
\item Associate weight $w_0^j = 1$ with each particle.
\end{itemize}

\noindent Then for each time step $\tau = 1, \ldots, t$:

\begin{itemize}[noitemsep,topsep=0pt]
\item Sample stochastic dynamic model to update each particle: $\textbf{x}_\tau^j \mid \textbf{x}_{\tau-1}^j \sim p(\textbf{x}_{\tau} \mid \textbf{x}_{\tau-1}, \theta)$
\item Scale each weight by the likelihood of the next observation: $w_\tau^j = w_{\tau-1}^j p(\textbf{y}_\tau \mid \textbf{x}_\tau^j, \theta)$
\item Optionally resample the particles $\{\textbf{x}_\tau^j\}$ to rebalance their weights $\{w_\tau^j\}$. We used Kitagawa's stratified scheme \citep{kitagawa1996stratified} on every time step. We preserve the sum of the weights through resampling because this represents the likelihood of the data observed so far: $p(\textbf{y}_{1:\tau}|\theta) \approx \sum_j w_\tau^j / N_x$
\end{itemize}

The important output here is the final estimate for the likelihood of the whole time series given the model parameters: $p(\textbf{y}_{1:t}|\theta) \approx \sum_j w_t^j / N_x$. For $N_x\geq1$, this is an unbiased estimate that could be summed across all sequences and used directly in a pseudo-marginal chain (with different random numbers for each PF run). Instead we operate with large numbers of particles (to reduce variance) and fix the random seed for each time series \citep{lee2008towards}, so that we can use the output as a deterministic function of $\theta$ in subset samplers. 

\section{Scalable Synthetic Task}
\label{app:synthetic}


The purpose of developing this sampling problem was to be able to run large comparisons for state spaces of more than 2 dimensions, without having to exercise a PF on every likelihood evaluation, but with noise characteristics that are similar to the disease model.

\begin{enumerate}[noitemsep]
\item Each data point (scenario $i$) has a single observed count rather than a time series $y_i \sim \text{Poisson}(z_i)$ where $z_i$ is not observable and would correspond to the occupancy of a compartment in the disease model.  
\item The non-observable latent state is itself drawn from a random process: $z_i \sim \text{Poisson}(\mu_i)$
\item The mean occupancy $\mu_i$ is a known function of the model parameters (state): $N_{pop} \sigma (\theta)^\intercal w_i $ where the sigmoid $\sigma(\theta) \equiv \frac{1}{1+\exp(-\theta)}$ transforms the unbounded $D$-dimensional state vector $\theta$ element-wise to $[0,1]^D$ and $w_i$ are projection weights that vary by scenario. These weights are used to control the extent to which different data points carry information about different dimensions of $\theta$. $N_{pop} \equiv 1250 D$ is the equivalent of the population size in the disease model.
\item The projection weights $w_i$ are each a sparse vector with 2 non-zero entries at locations $j$ and $(j+1) \mod D$ where $j$ is a uniform random offset. The values of these non-zero entries are $u_i$ and $1-u_i$ respectively, where $u_i \sim \text{Uniform}[0,1]$.
\item We optionally include an extra correlation between each consecutive pair of dimensions of $w_i$ in proportions $(1-\zeta, \zeta)$ with $\zeta = 0.45$ (for \Cref{subsec:corr}) and 0.485 (for \Cref{subsec:breakdown}). This has the effect of introducing ambiguity between those dimensions, causing the likelihood to have a ridged structure (making large random walk proposals less likely to be successful). In the latter `extreme' correlation case we include a spherical Normal prior located at the origin to constrain the target density within a plausible region; the variance of this prior is 25 for regular data and 9 for `tall' data.
\end{enumerate}
The true likelihood function $l(y_i|\theta, w_i)$ for an observed data point $y_i$ given the candidate model parameters $\theta$ and the known context $w_i$ involves taking the expectation over the hidden $z_i$. We average the likelihood over $n_{reps} = 16$ realisations of $z_i$. This is analogous to the average over particles in the PF likelihood calculation. By varying $n_{reps}$ we trade-off computational cost with the noise level in this estimate.

The random seeds used for these realisations of the hidden state must be fixed (for any given state) to make the likelihood evaluation repeatable, at the cost of some per-scenario bias. As the interpretation of random numbers in a PF can cause the likelihood bias to change with small changes in $\theta$, the same effect is achieved here with random seeds that are only constant within cells of a state space grid of resolution $10^{-3}$ in $\log(1+\exp \theta)$. The phase offset of the grids are different for each of the repetitions.

Versions of this test problem were created with dimension D = {2, 4, 16} and a simplified version with $D=1$ was use to illustrate \Cref{subsec:Motivation}. A `smooth' variant used as an easier test for all samplers (to verify adaptive step size benefits) was obtained by using the distribution mean $\mu_i$ in place of $z_i$. This is biased compared with full marginalisation over $z_i$ and so is only useful as a test. The computation budget was 8192 full likelihood evaluations for the main tasks but 4096 for the smooth variant.

A `tall data' variant increased the data set size so that each of the 64 likelihoods contained a batch of 250 data points instead of a single item. Therefore, the full effective data set size was 16,000. Each data point was chosen to carry less information and less noise by making  $z_i \sim \text{Poisson}(10^{-4}\mu_i)$ and $N_{pop} = 10^5$. This configuration was chosen to artificially allow a smaller number of Poisson draws, $n_{reps} = 4$, to limit computational demands. Additionally, a smaller computation budget was used (2048 full evaluations). In the context of disease modelling, these parameters might be reasonable for data collected over small time windows or small sub-populations, such that each positive count is very small and conveys little information on its own.

\section{Performance Metrics}
\label{app:performance_metrics}

We compute a number of different metrics to understand the mixing performance and asymptotic accuracy of each sampler, and to provide insight on how they each make use of the computational budget in different ways. All tabulated metrics were computed on the second half of each sampling run, in terms of computation. For `strict' exact samplers, proxies and adaptation mechanisms were frozen for the measurement interval. 

The confidence intervals shown in the tables are for the median of each metric across 50 runs. The reason for reporting median rather than mean is to ignore outliers: a few divergent sampling runs usually do not damage a combined sample. The confidence intervals represent the 5th and 95th percentiles (obtained by bootstrapping); non-overlapping intervals between any two samplers are statistical evidence for a difference. For the metrics below, we define:
\begin{itemize}[noitemsep,topsep=0pt]
    \item $d$ is the dimensionality of the task's state space such that $\theta \in  \mathbb{R}^d$;
    \item $N$ is the number of scenarios in a full likelihood evaluation;
    \item \(\sigma_0^2\) is the square of the task's default proposal size;
    \item $T$ is the total number of samples in the measurement interval;
    \item $(\theta_t)_{t=0}^{T}$ are the states output by the chain inclusive of initial state $\theta_0$;
    \item $C$ is the number of single-scenario likelihood evaluations in the interval;
    \item $a_t$ is the calculated acceptance probability for the top-level chain at step $t$.
\end{itemize}

\subsection{Kullback-Leibler Divergence}
\label{app:metric-KL}


We report an absolute measure of sampling accuracy by computing the KL divergence between Multivariate Normal fits for the sample under consideration ($\mathcal{N}_1$) and a `reference' sample ($\mathcal{N}_2$): 

\label{app:kl_divergence}
\begin{align*}
D_{\text{KL}}(\mathcal{N}_1 \| \mathcal{N}_2) = \frac{1}{2} \Bigg[ 
& \operatorname{Tr}(\boldsymbol{\Sigma}_2^{-1} \boldsymbol{\Sigma}_1) + 
(\boldsymbol{\mu}_2 - \boldsymbol{\mu}_1)^\top \boldsymbol{\Sigma}_2^{-1} (\boldsymbol{\mu}_2 - \boldsymbol{\mu}_1) - d + \log\left(\frac{\det \boldsymbol{\Sigma}_2}{\det \boldsymbol{\Sigma}_1}\right) 
\Bigg],
\end{align*}

\noindent where $\boldsymbol{\mu}, \boldsymbol{\Sigma}$ represent sample mean and covariance in each case. 

We obtained a good quality reference sample for each task by running 16 parallel adaptive MCMC chains, each with a computation budget 8 times larger than we use in the comparison of samplers. For tasks with synthetic data, we start each chain at the known `true' state.

This metric would not be available when working with new data sets, because the true model parameters would be unknown, but we regard it as an essential objective measure of performance for \emph{comparing samplers}. A sample will only achieve a low error according to this metric if it accurately captures the mean and the dispersion (in the form of covariance) across all state dimensions.

\subsection{Computation per Step}
\label{app:metric-comp}
The total computation $C$ is divided by number of scenarios $N$ to obtain the number of full `sweeps' of the data set in the measurement interval. This is then divided by the number of steps (iterations) $T$ to obtain full evaluations per step:

$$\textrm{evals/step} = \frac{C}{NT}.$$

This metric is approximately 1 for a simple MCMC chain because each new proposal must be evaluated on all scenarios (all terms of full target density). Subset samplers typically achieve $<1$ on this metric.

\subsection{Acceptance Probability \& Acceptances per Evaluation}
\label{app:metric-accept}

The acceptance probability is an average for the top-level chain of each sampler. Rather than count the accept or reject outcomes (in $\{0,1\}$), variance is reduced by averaging the known acceptance probabilities $a_t \in [0,1]$ used as inputs to each acceptance rule. Where this is not computed directly in the sampler, we calculate it. For HINTS, a zero-proposal ($\theta' = \theta$) may occur at the root-node; we count this as a rejection.
$$\textrm{acceptance probability} = \frac{1}{T} \sum_{t=1}^T a_t$$
We report the acceptance probability as a percentage in results tables. When the computational cost is variable, it is more useful to have an acceptance metric that takes this into account. Therefore, we define acceptances per evaluation:
$$\textrm{accept/eval} = \frac{N}{C}  \sum_{t=1}^T a_t.$$
For a simple MCMC chain, $C \approx NT$, so this acceptance rate is the same as the acceptance probability.

\subsection{Variance per Evaluation}
\label{app:variance_rate}
This is the measure of mixing performance directly maximised by the cost-aware adaptation:
\begin{align}
    \textrm{variance/eval} = \frac{N}{C\sigma_0^2} \sum_{t=1}^{T} \| \theta_t - \theta_{t-1} \|^2.
\end{align}

%
\subsection{Effective Sample Size per Evaluation}
\label{app:metric-ESS}
The integrated autocorrelation time (IACT), defined as $\tau$, is used to quantify the correlation between MCMC samples. It is defined as 
$$\tau = 1 + 2 \sum_{i=1}^{P} \rho_{i},$$
where $\rho_i$ is the autocorrelation function (ACF) at lag $i$ of $(\theta_t)_{t=0}^{T}$, and $P$ is a cut-off period which serves to exclude the noisy estimates for large lags, obtained automatically \citep{sokal1997monte}. 
An Effective Sample Size (ESS) can then be calculated as $T/\tau$. To account for computational costs, the effective sample size per full evaluation is:

$$\textrm{ESS/eval} = \frac{N T}{C \tau}$$

We calculate the ESS metrics for each state dimension separately and report the minimum (worst) value across dimensions.

\subsection{Gelman-Rubin Statistic ($\hat{R}$)}\label{app:Rhat}

The Gelman-Rubin diagnostic \citep{gelman1992inference} measures the convergence of multiple MCMC chains. It compares the between-chain variance to the within-chain variance to estimate how well the chains have mixed. 

The within-chain variance $W$ is the mean square Euclidean distance of each state encountered from the mean for the individual chain. The between chain variance $B$ is the mean square Euclidean distance from the common mean across all chains. Then a useful measure of convergence is:
$$\hat{R} = \sqrt{\frac{W+B}{W}}.$$

For successful sampling, $\hat{R}$ should approach 1, and \Citet{carpenter2017stan} suggest a value lower than 1.05 signifies that all chains have likely converged to the target distribution. Larger values indicate that some or all chains have failed to converge. This is the only metric that measures consistency between chains directly.

\section{HINTS is an Exact Sampler}
\label{app:proof}

A proof is now given that HINTS is an exact sampler for the full target density in \Cref{eq:target}. Consider the top-level HINTS chain, at the root-node $h=H$, $j=1$, defined in \Cref{subsubsec:HINTS}. \Cref{alg:HINTS} constructs a directed (asymmetric) proposal for a MH chain running at this level. Exact sampling of target density $\pi$ is achieved if the chain's transition function $t(\theta,\theta')$ satisfies the detailed balance property:

\begin{equation} \label{eq:BALANCEROOT} \frac{t(\theta,\theta')}{t(\theta',\theta)} = \frac{\pi(\theta')}{\pi(\theta)}. \end{equation}

For simplicity of notation in reasoning about a specific node (level $h$, index $j$) \emph{locally} we drop these labels from  the subset information function, $F(\theta)$. For the root-node \emph{only}, HINTS requires $F(\theta)$ to be proportional to the target density, so  \Cref{eq:BALANCEROOT} is equivalent to:
\begin{equation} \label{eq:BALANCE} 
\frac{t(\theta,\theta')}{t(\theta',\theta)} = \frac{F(\theta')}{F(\theta)}. \end{equation}

Below the root ($h<H$), the subset information function is not limited to likelihoods; our result applies equally when a fitted proxy or cheap-to-compute approximation is substituted for an expensive likelihood evaluation. 

Our proof follows the same idea as \cite{Hastings1970}: an asymmetric proposal constructed by an embedded process can be used in an MCMC sampling chain if the asymmetry is known and accounted for. The additional complexity arises from the fact that the HINTS sampling process holds state at multiple levels in the hierarchy. Furthermore, our \textit{Hastings' correction} is path-specific rather than representing the whole proposal density. We construct a proof by induction of \Cref{eq:BALANCE} for every node in the tree, and therefore obtain detailed balance at the root.

\Cref{subsec:CMHPROOF} shows that the HINTS acceptance rule satisfies \Cref{eq:BALANCE} for every node in the tree, if a particular ratio (between the actual forward trajectory probability and the notional reverse trajectory probability) is supplied by each embedded (child) process. This replaces the conventional asymmetry correction constructed from a known proposal density.

\Cref{subsec:INDSTEP} shows that HINTS does indeed compute this ratio at every non-root-node ($h<H$) in the tree (given that the balance property holds at that level), therefore establishing the balance property at the parent node (level $h+1$). In \Cref{subsec:IND} the balance property is shown to hold at leaf nodes ($h=0$). Then the required result is obtained by induction over the whole tree. 

\subsection{Detailed Balance Given Trajectory Probability Ratios\label{subsec:CMHPROOF}}

Let $\omega$ represent the sequence of states (`trajectory' or `path') of a lower-level process, drawn from a set $\Omega$. With continuous-valued $\theta$, this is an infinite set, but our result also holds for discrete state spaces. Let $\Omega(\theta,\theta')$ be the subset of paths in $\Omega$ that start at state $\theta$ and end at state $\theta'$. Also let $\bar \omega$ be the \emph{reverse path}. The mapping between forward and reverse paths is required to be a bijection satisfying:
\begin{equation} \omega \in \Omega(\theta,\theta') \iff \bar{\omega} \in \Omega(\theta',\theta). \end{equation}

Let $p(\omega|\theta)$ be the probability density for paths generated by the lower-level process that constructs proposals for the current node, given start state $\theta$. The ratio $p(\bar \omega|\theta)/p(\omega|\theta')$ is supplied by the lower-level process. HINTS applies a MH rule to accept or reject the composite proposal $\theta'$. This is the same as conventional MH but with $p(\bar\omega|\theta')/p(\omega|\theta)$ substituted for $q(\theta|\theta')/q(\theta'|\theta)$. Therefore, the acceptance probability is $1 \wedge \phi(\omega)$ where:
\begin{equation} \label{eq:CMH} \phi(\omega) = \frac{F(\theta')p(\bar\omega|\theta')}{F(\theta)p(\omega|\theta)} .\end{equation}
Consider first the LHS of \Cref{eq:BALANCE} for the case $\theta' \neq \theta$. A change of state can only result from an acceptance, and therefore we can obtain the total transition probability by integration of the acceptance probability over the paths ending in $\theta'$:
\begin{equation} \label{eq:TRAT}
\frac {t(\theta,\theta')}{t(\theta',\theta)} = \frac{\E_{\{\omega \in \Omega(\theta,\theta'), \omega \sim  p(\omega|\theta)\}}
 (1 \wedge \phi(\omega))}{\E_{\{\omega' \in
\Omega(\theta',\theta), \omega' \sim p(\omega'|\theta') \}} (1 \wedge \phi(\omega'))}.
\end{equation}

The bijection between forward and reverse paths allows each specific choice of $\omega$ in the numerator to be paired with the choice $\omega' = \bar \omega$ in the denominator. Noting the identity $\phi(\bar \omega) = 1/\phi(\omega)$, obtained by inspection of \Cref{eq:CMH}, the ratio of these two terms is:
\begin{equation} \label{eq:RESULT} \frac{p(\omega|\theta) (1 \wedge \phi(\omega))}{p(\bar \omega|\theta') (1 \wedge 1/\phi(\omega))} =
\frac{p(\omega|\theta)}{p(\bar \omega|\theta')} \phi(\omega) =
\frac{F(\theta')}{F(\theta)}.\end{equation}
As \Cref{eq:RESULT} holds for every such pair $(\omega,\bar \omega)$, the ratio of expectations is also $\frac{F(\theta')}{F(\theta)}$ and the balance property \Cref{eq:BALANCE} is proven for the case $\theta' \neq \theta$. The proof for $\theta' = \theta$ is trivial, by substitution.

\subsection{Inductive Step\label{subsec:INDSTEP}}

To apply this result to HINTS, let a path $\omega$ consist of the series of states $(\theta_0,\theta_1,\ldots,\theta_m)$ experienced under a sequence of transitions at level $h$ using the associated subset information functions. (Where a rejection has taken place, consecutive elements will be identical.)  For every such path there exists exactly one reverse path $\bar \omega$, in which the sequence is reversed, assuming the scenarios are shuffled according to \Cref{subsubsec:HINTS}. Therefore, we have the required bijection between forward and reverse paths. The probability of experiencing each forward path is the product over individual transitions:
\begin{equation} \label{eq:FWD}
p(\omega|\theta) = \prod_{i= 1, \ldots,m} t(\theta_{i-1},\theta_i).
\end{equation}
The same quantity can be computed for the notional reverse path:
\begin{equation} \label{eq:REV}
p(\bar \omega|\theta') = \prod_{j = 1, \ldots, m} t(\theta_{m+1-j},\theta_{m-j}).
\end{equation}
Rewriting \Cref{eq:REV} using  $k = m + 1 - j$ yields:
\begin{equation} \label{eq:REV2}
p(\bar \omega|\theta') = \prod_{k = 1, \ldots, m} t(\theta_k,\theta_{k-1}).
\end{equation}
Combining \Cref{eq:FWD} and \Cref{eq:REV2}:
\begin{equation} \label{eq:TRAT2}
\frac{p(\omega|\theta)}{p(\bar \omega|\theta')} = \prod_{k= 1, \ldots,m}
\frac{t(\theta_{k-1},\theta_k)}{t(\theta_k,\theta_{k-1})}.
\end{equation}
Using \Cref{eq:BALANCE}, at level $h$, for each term in the product yields:
\begin{equation} \label{eq:BALANCEHINTS}
\frac{p(\omega|\theta)}{p(\bar \omega|\theta')} = \prod_{k= 1, \ldots,m} \frac{F_k(\theta_k)}{F_k(\theta_{k-1})},
\end{equation}
%


\noindent where $F_k$ is the subset information function for child $k$ of the current node. The RHS of \Cref{eq:BALANCEHINTS} is accumulated by the algorithm as a product of known quantities from the lower-level transitions. Therefore, the new rule is applied correctly and the balance property is obtained at level $h+1$.

\subsection{Proof by Induction \label{subsec:IND}}

At leaves of the HINTS tree, primitive transitions are given by application of the MH rule using an appropriate fixed proposal density, and therefore \Cref{eq:BALANCE} holds directly from the existing results. The result in \Cref{subsec:INDSTEP} can then be applied inductively for each node at level $h = 1, \ldots, H$ to obtain detailed balance of the root-node MH chain.

To guarantee that the root-node generates true samples from the target density, it remains to ensure that the process is ergodic and aperiodic. In regular MH chains, these properties can be achieved by positive-everywhere proposals such as a Gaussian random walk kernel, and many other proposal functions. In HINTS, combining positive-everywhere leaf node proposals with positive-everywhere subset-information functions for all levels $j$ and nodes $h<H$, is one way to ensure the composite proposals experienced by the root are also positive-everywhere. $\blacksquare$


\vskip 0.2in

\bibliographystyle{abbrvnat}
\bibliography{references.bib}

\end{document}